%% file: manuscript.tex
\documentclass[manuscript,screen, acmsmall]{acmart}

\usepackage{multirow}
\usepackage{multicol}
\usepackage{comment}
\usepackage{subfig}
\usepackage{soul}
\usepackage{array} % <-- new
\usepackage{tabularx}
\usepackage{makecell}
\newcolumntype{C}[1]{>{\centering\arraybackslash}p{#1}} 
\usepackage[normalem]{ulem}

\AtBeginDocument{%
  \providecommand\BibTeX{{%
    \normalfont B\kern-0.5em{\scshape i\kern-0.25em b}\kern-0.8em\TeX}}}

\copyrightyear{2025}
\acmYear{2025}
\acmDOI{XXXXXXX.XXXXXXX}

\setcopyright{acmlicensed}
\begin{document}

% \title{Crowded Social Sensing: Analyzing Noisy Social Interactions through Smartwatch Audio and Motion Data}

\title{Detecting In-Person Conversations in Noisy Real-World Environments with Smartwatch Audio and Motion Sensing}


\author{Alice Zhang}
\email{alice.zhang@austin.utexas.edu}
\orcid{0009-0006-0148-095X}

\author{Callihan Bertley}
\email{calbertley@utexas.edu}

\author{Dawei Liang}
\email{dawei.liang@utexas.edu}

\author{Edison Thomaz}
\email{ethomaz@utexas.edu}
\affiliation{%
  \institution{The University of Texas at Austin}
  \city{Austin}
  \state{Texas}
  \country{USA}
}

\renewcommand{\shortauthors}{Zhang et al.}

\begin{abstract}
  Social interactions play a crucial role in shaping human behavior, relationships, and societies. It encompasses various forms of communication, such as verbal conversation, non-verbal gestures, facial expressions, and body language. In this work, we develop a novel computational approach to detect face-to-face verbal conversations, a foundational aspect of human social interactions. We leverage multimodal data captured by a commodity smartwatch, specifically synchronizing microphone audio with 6-axis inertial signals (accelerometer and gyroscope). We design, train, and evaluate convolutional and attention-based neural networks using three different fusion methods to integrate the audio and motion modalities. To validate this framework, we conduct a \textit{lab} study with 11 participants and a \textit{semi-naturalistic} study with 24 participants. Our comprehensive evaluation demonstrates that fusing inertial data with audio significantly improves detection performance by capturing non-verbal conversational dynamics. Overall, our framework achieved 82.0$\pm$3.0\% macro F1-score when detecting conversations in the lab and 77.2$\pm$1.8\% in the semi-naturalistic setting. Lastly, we demonstrate real-time conversation detection by deploying our trained model to a user application running on a commercial smartwatch.
  % Abstract should be improved. "foundational aspect of human social interactions - which ones ? Which data are captured, used in relation to smartwatch. You "analyzed deep learning models" (which), you trained them also or not ? Which inertial data you used ? A lot of evaluation setting and results, while it is not clear how complete setup was developed, which data were used and how.

\end{abstract}




\maketitle
\begin{figure}
    \centering
    \includegraphics[width=0.9\linewidth]{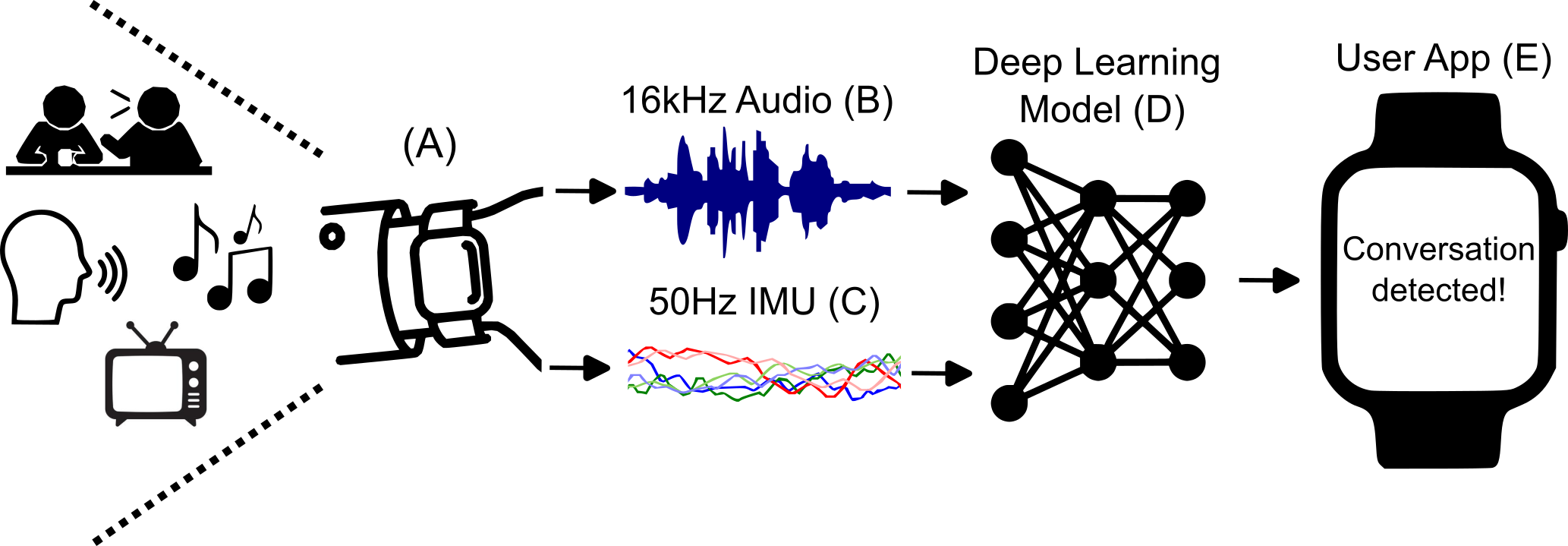}
    \caption{For conversation detection in noisy environments, we use a wrist-worn smartwatch (A) to capture 16kHz audio (B) and 50Hz IMU motion (C) data. The two sensor data streams feed into a multimodal deep learning model (D) to detect instances of conversation in real-time in a user application (E).}
    \label{fig:pipeline_diagram}
    \vspace{-10pt}
\end{figure}
\section{Introduction}
\input{intro}

\section{Related Work}
\input{related_work}

\section{Conversation Modeling}
\input{conversation_modeling}

\section{Data Collection}
\input{data_collection}

\section{Social Event Detection Methodology}
\input{social_event_detection}

\section{Evaluation and Results}
\input{eval_and_results}

\section{Discussion}
\input{discussion}

%\section{Model Deployment}
%\input{deployment}

\section{Applications}
\input{application}

\section{Limitations and Future Work}
\input{limitations_and_future_work}

\section{Conclusion}
\input{conclusion}

\section{Code Availability}
The code used in this study is not currently publicly available due to a pending patent application. It is available from the corresponding author upon reasonable request, subject to appropriate intellectual property and confidentiality considerations.
\begin{comment}
\section{Insecurities}
- you're putting reading out loud into the same class as people not participating in a group discussion. The potential issue is that reading out loud involves holding a book so there is a very distinct inertial (or lack thereof) pattern with reading out loud while holding a book compared to someone not participating in a group discussion. Also the narrative is that other speech is an important class especially for team dynamics but not speaking and speaking too much (monologue or rambling on) are in the same class.
- um oh dear should the this actually be a four class problem? Background noise, dominating the conversation (monologue?), not speaking in conversation, conversation? Or should you investigate this class within the “other speech” class? Since you're especially focused on the "other speech" class?
- and um should you have done a frame sesitivity analysis of the 30s conversation window? especially if these are busy environments with rapidly changing dynamics
\end{comment}

\bibliographystyle{ACM-Reference-Format}
\bibliography{references}

\end{document}

%% file: intro.tex
% You talk about social interactions in general, importance, other methods to sense and monitor social interactions in real world settings. It is not sufficient to elaborate what are their main issues and limitations. What has to be improved etc. You then propose a new computational direction (platform) regarding social interaction sensing, where you mention that you additionally capture non-verbal communication gestures - this is wide description ... you do not elaborate which one exactly to take into consideration. It is then hard to understand, how some smartwatch and capture data can help with this. 

Social interactions play a crucial role in shaping human behavior, relationships, and societies. It encompasses various forms of communication, such as verbal conversation, non-verbal gestures, facial expressions, and body language. Critically, lack of social interaction and loneliness are globally growing public health concerns, especially following the COVID-19 pandemic \cite{who2021social}. Prior research has shown that social isolation and loneliness are comparable to well-established risks for premature mortality, such as obesity and substance abuse \cite{pantell2013social}. 

Conventional methods of assessing an individual’s social connections rely on retrospective clinician-rating surveys \cite{reis1991studying} or momentary self-report questionnaires called ecological momentary assessments (EMAs) \cite{moskowitz2006ecological}. However, these methods have many shortcomings; they are susceptible to recall biases and often impose a significant burden on individuals. Moreover, these methods may not be appropriate for individuals who suffer from communication disorders, such as those with cognitive or language impairments. While recent passive sensing approaches using audio have attempted to automate this process, they face critical challenges in real-world deployment. Audio-only models often degrade significantly in noisy environments and fail to distinguish between the user’s participation and surrounding background chatter \cite{liang2023automated}. Furthermore, they ignore body language, the physical dimension of human communication which offers vital context often lost in audio processing alone. Consequently, a passive and universally-accessible method to sense and monitor social interactions in real-world settings would address these limitations.

In this work, we leverage the rich sensing capabilities of smartwatches. These devices are compelling for human-centered applications since they are generally comfortable to wear, socially acceptable, and increasingly ubiquitous. Importantly, they do not carry the stigma or burden of bulkier, customized sensors. Using smartwatches, we advance a new computational direction in social interaction sensing in real-world environments. We combine acoustic and inertial data to capture two key aspects of human communication: \textit{in-person verbal conversations} and \textit{conversational gesticulation} (hand and arm gestures accompanying speech) (Figure \ref{fig:pipeline_diagram}). We then deploy a joint acoustic-inertial deep learning model to a smartwatch to detect conversations in real-time.

Through our multimodal approach, we capture hand movements that naturally accompany speech (gesticulation) to emphasize or illustrate speech \cite{CloughDuff2020Gesture}. Leveraging non-verbal communication gestures is highly relevant in social interaction sensing, since body movements in conversations can provide as much information as the spoken content itself \cite{mehrabian1972, bavelas2007conversational}. By utilizing the accelerometer and gyroscope sensors on a smartwatch, we employ the inertial patterns of these wrist movements to distinguish active conversation from passive listening or background noise. Prior work focused on social interaction analysis has not considered this important dimension of social interaction sensing \cite{liang2023automated, white2023socialbit, zhang2016dopenc, cattuto2010dyanmics, luo2013socialweaver, miluzzo2010darwin}. Additionally, we emphasize the analysis of social activities in acoustically challenging environments aligned with real-world settings where an audio-only approach may decline in performance with noise. 

We formulate the detection task around conversational turn-taking. While social interaction modeling approaches vary, turn-taking is widely agreed to be a universal characteristic of conversation across languages and cultures \cite{holler2016turn, donaldson1979one, sacks1975simple}. Therefore, we approach the task as a three-class classification problem, distinguishing engaged \textit{conversation} from \textit{other speech} (e.g., monologues) and \textit{background noise}. This task formulation allows us to specifically isolate social interactions where the user is an active participant rather than a passive observer.

The specific contributions of this work are:

\begin{enumerate}

    \item A novel multimodal approach to demonstrate the robustness and reliability of conversation detection in noisy and dynamic environments with acoustic noise levels ranging between 50-70dBA. %The approach leverages the acoustic and inertial sensors of a commodity smartwatch to capture both verbal and non-verbal conversation behaviors. Our method is also speaker agnostic and does not need to be customized to specific individuals.
    
    \item An extensive evaluation of various modeling techniques and fusion methods showcasing the effectiveness of combining audio and inertial modalities for conversation detection, especially in environments with significant acoustic noise. We illustrate how incorporating wrist-motion data clarifies situations that are confused by a single modality alone, with our framework achieving 82.0$\pm$3.0\% and 77.2$\pm$1.8\% macro F1-scores for detecting conversations in \textit{lab} and \textit{semi-naturalistic} studies respectively. 

    \item A thorough set of analyses to illustrate the benefits of multimodal sensing across multiple contexts and audio sampling rates. To increase privacy protection in acoustic sensing, we show that the addition of inertial sensing to capture non-verbal conversational gestures can effectively supplement information lost in downsampled audio.

    \item A demonstration of the feasibility of our multimodal real-time sensing approach for smartwatches. We demonstrate that despite the model complexity, we can optimize and deploy the joint acoustic-inertial model onto a single commodity smartwatch with an average inference time less than 1s. We also perform a rigorous cost-benefit analysis of the smartwatch battery life and model performance across various sampling rates of our multimodal sensing method.

    \item A new, annotated dataset of social activities from 35 participants split into 11 groups across \textit{lab} and \textit{semi-naturalistic} settings with synchronized audio features and raw inertial data collected using an off-the-shelf smartwatch. This dataset enables research in disciplines ranging from understanding non-verbal communication in busy social contexts to improving technology interfaces. The data is available \href{https://osf.io/hmy3n/?view_only=f11b0d43a5a54f54bcffe1ee0adcff63}{here}.
    
\end{enumerate}

%In face-to-face communications, non-verbal communication including gestures, facial expressions, posture, and eye contact, communicates as much information as the spoken content itself \cite{mehrabian1972, bavelas2007conversational}.

% While prior works have used the sensing capabilities of smartwatches for human activity recognition (HAR) \cite{bhattacharya2022leveraging, becker2019gestear, chun2019towards, chun2020eating, weiss2016smartwatch}, our work focuses exclusively on the detection of conversations.

%% file: related_work.tex
%You start with Smartwatch Multimodal Sensing. Since you talk about before about new computational platform for social interaction sensing, I would expect here that you will talk about other systems, and how your system relates to them, with new solutions, ideas, and by using e. g. smartwatch. Especially in relation to non-verbal information. Further, how accurate and in real-time capturing is among different systems. I would advise that you rewrite this section. 

\subsection{Social Interaction Sensing}

Existing methods for sensing face-to-face social interactions have primarily relied on smartphone-based or wearable sensor-based methods.

\subsubsection{Smartphone-Based Methods}

Smartphone-based methods utilize sensors in smartphones to detect and analyze social interactions. SocialWeaver \cite{luo2013socialweaver} and DopEnc \cite{zhang2016dopenc} use Bluetooth and doppler profiling, respectively, with signals transmitted and received between individuals' smartphones to determine the proximity between two individuals and infer whether they are engaged in conversation. Similarly,  Crowd++ \cite{xu2013crowd} uses audio collected from smartphones to estimate the number of speakers in a group. However, these smartphone-based methods do not work as intended when individuals do not carry their smartphones on body but instead in a purse or backpack, as is commonly observed in some populations when not actively using the phone \cite{redmayne2017where}. Therefore, the assumption of these methods that smartphones are primarily carried on-body limits these methods' scope of application.

There are also collaborative sensing methods involving multiple smartphones for social interaction analysis, such as SocioPhone \cite{lee2013sociophone} and Darwin \cite{miluzzo2010darwin}, which use individuals’ voice signals captured by microphones across multiple smartphones. Data from multiple devices are shared to detect conversational turns as in SocioPhone or perform speaker recognition as in Darwin. Additionally, Li \textit{et al.} leverages a smartphone in coordination with multiple on-body inertial sensors to monitor in-person interactions \cite{li2013multimodal}. However, employing multiple devices can restrict their ease of scalability.

\subsubsection{Wearable Sensor-Based Methods}

An alternative to smartphone sensing is using custom or commercial wearable sensors to infer interactions among individuals. Previous studies have developed custom hardware devices, such as sociometric badges with infrared sensors (IR) \cite{oguin2009sensible} or active radio frequency identification (RFID) tags \cite{cattuto2010dyanmics}, worn around the neck with a lanyard. Both the IR sensor and RFID tag identify instances when two individuals are directly facing each other within close proximity (<1 meter), indicating that the individuals are engaged in an interaction. However, these wearable badge-based methods require each individual in the interaction to have and to wear their own badge. Hence, these methods are not easily scalable to social interaction sensing in general populations. 

Rahman \textit{et al.} \cite{rahman2011mconverse} and Bari \textit{et al.} \cite{bari2018rconverse} both developed conversation detection methods from respiration signals collected by a chest band worn around a speaker's chest.  Bari \textit{et al.} further leveraged electrocardiogram sensors within the chest band and inertial sensors on a wristband worn on the dominant hand to specifically detect stressful conversations. However, continuously wearing a chest band for everyday sensing can be inconvenient and interfere with daily activities. 

%Recent works have advanced towards using a single device on one user to model spoken interactions between the user and other individuals. For instance, Ghahramani \textit{et al}. \cite{ghahramani2018learning} hypothesized that social interactions between co-workers in an office building impact ambient conditions, such as sound and CO2 levels. Consequently, they created a wearable sensor box to monitor ambient conditions to detect whether office building occupants are working with others or working alone. 

Most recently, commodity devices have been used to recognize face-to-face social conversations. Commercially, Apple AirPods Pro feature a Conversation Awareness mode that automatically reduces media volume and background noise upon detecting the user’s speech \cite{airpods}. Off-the-shelf smartwatches have also been used to detect and quantify face-to-face social interactions. In studies by Liang \textit{et al}. \cite{liang2023automated}, White \textit{et a}l. \cite{white2023socialbit}, and Ahmed \textit{et al.}\cite{ahmed2025socialpulse}, the microphone of a commodity smartwatch captures acoustic features that are used to detect instances of in-person conversations. Additionally, White \textit{et al.} specifically requires collecting voice samples and developing a voiceprint unique to each user during model training in order to compare the input audio to the pool of known speaker identities during model inference. 

Our work differs from these prior works in that it leverages multiple sensors within a single commodity smartwatch to infer instances of face-to-face conversations. Unlike White \textit{et al.} which requires customizing the model to specific users, our framework is speaker agnostic \cite{white2023socialbit}. This reduces the required pre-training overhead and increases user privacy as we do not need to collect voice samples for speaker identification or verification - voice samples that if misused or leaked could be leveraged for voice spoofing.

Additionally, a major limitation of prior works is their reliance on single-modality sensing (typically audio). With the exception of Li \textit{et al.} \cite{li2013multimodal}, who incorporate on-body inertial sensors, prior systems largely overlook non-verbal cues. Non-verbal cues are essential for robust conversation detection, especially in loud or busy environments where other models struggle to distinguish between active participation and background noise. %The current work also seeks to improve model performance in especially loud and busy environments, such as restaurants or bars, where background conversations can be confused for a device user's conversations.

Lastly, many existing approaches rely on offline data processing \cite{ cattuto2010dyanmics, oguin2009sensible, li2013multimodal, xu2013crowd}. For audio-systems, this is especially limiting for practical applications due to the privacy risks associated with storing raw speech on-device prior to processing. To address this, our method processes only the most recent 30-second buffer and immediately discards the raw audio following inference, which is a process that takes less than one second on-device. Conversely, while real-time approaches exist  \cite{rahman2011mconverse, liang2023automated, airpods, lee2013sociophone, miluzzo2010darwin, white2023socialbit, ahmed2025socialpulse}, they remain primarily audio-based and consequently suffer from the robustness limitations discussed above.
%real-time: Liang \textit{et al}., Apple AirPods Pro, Bari \textit{et al.}, \cite{lee2013sociophone, miluzzo2010darwin} 
%offline: White \textit{et al.}, \cite{cattuto2010dyanmics, oguin2009sensible, luo2013socialweaver, xu2013crowd, li2013multimodal}
%DopEnc - unknown

\subsection{Smartwatch Multimodal Sensing}

The combined use of audio and inertial sensing with commodity smartwatches has been broadly explored in Human Activity Recognition (HAR) and Human-Computer Interaction (HCI) applications and present a unique opportunity to capture the non-verbal gestures missing from the systems described in the previous section. Kim \textit{et al.} used accelerometer and acoustic signals to classify 5 daily activities including eating, vacuuming, sleeping, showering, and watching TV \cite{hyunchoong2017collaborative}. Similarly, GestEar \cite{becker2019gestear} jointly leveraged accelerometer, gyroscope, and acoustic signals to perform gesture classification on a limited set of simple gestures centered around snapping, knocking, and clapping. 

Towards using audio and inertial sensing to recognize a greater number of activities, Siddiqui and Chan collected a dataset from inertial sensors and microphones as pressure-based sensors placed at the wrist to recognize a set of 13 daily life gestures and 1 relaxed gesture  \cite{siddiqui2020multimodal}. From this data, they hand-crafted and selected the most relevant features to gesture recognition using a mutual information-based algorithm that were then fed as inputs to classical machine learning models. Moreover, the ExtraSensory Dataset, collected from both smartwatches and smartphones of 60 participants, contains an even wider set of activities with user-labeled contexts \cite{vaizman2018extrasensory}. The dataset contains IMU, location, phone state, and phone-recorded audio for classification of user contexts and activities, such as "at school" or "driving".

Additional related work comes from Mollyn \textit{et al.} who presented SAMoSa, a framework for recognition of 26 daily activities across several indoor environments using inertial signals and downsampled audio \cite{mollyn2022samosa}. Likewise, Bhattacharya \textit{et al.} collected synchronized audio and inertial data from a smartwatch for a set of 23 activities, such as writing and typing, for ADL recognition in \textit{semi-naturalistic} and \textit{in-the-wild} environments \cite{bhattacharya2022leveraging}.  Liang \textit{et al.} further demonstrated a teacher-student framework to build an IMU-based HAR model for greater accuracy in recognizing this set of activities for ADL recognition \cite{liang2022audioimu}. During training, the IMU model is augmented with acoustic knowledge, and once trained, the model only uses motion inputs for inference. 

Smartwatch sensing has increasingly been applied to social interaction sensing; however, prior efforts have been limited to audio-only approaches. SocialBit \cite{white2023socialbit} and Liang \textit{et al.} \cite{liang2023automated} leverage smartwatch microphones to detect face-to-face social interactions. Furthermore, SocialPulse \cite{ahmed2025socialpulse} introduces the capability of detecting virtual social interactions using smartwatch acoustic features. While all three methods implement real-time sensing, joint acoustic-inertial sensing to capture multimodal dynamics of social interactions remains unexplored.

Contrasting against this previous body of research on smartwatch multimodal sensing, our work focuses exclusively on the detection of in-person conversations in challenging real-world scenarios. We leverage inertial data to capture non-verbal behaviors from in-person conversations to aid in conversation detection when the audio modality alone is otherwise insufficient for the task due to background sounds. Another difference is that in prior work, participants were instructed to wear the smartwatch on a specific hand (usually dominant hand) to better capture hand-based motion patterns of activities \cite{mollyn2022samosa, liang2022audioimu, hyunchoong2017collaborative, becker2019gestear, bari2018rconverse}. In our studies, however, participants were free to choose the hand on which to wear the smartwatch.
%\textcolor{blue}{In contrast to existing social interaction sensing methods via smartwatches that only use audio sensing \cite{liang2023automated, white2023socialbit, ahmed2025socialpulse}, we are the first to use synchronized audio and inertial data.} 
\subsection{Speech Processing Tools}

Speech processing tools have recently grown significantly in their capabilities. Speech processing tasks include speaker diarization, speaker recognition and verification, speech recognition, and more. Whisper, for instance, is an automatic speech recognition (ASR) system that enables real-time transcription and translation in multiple languages \cite{radford2022whisper, whispercpp}.  \textit{pyannote.audio} is a speaker diarization pipeline that recognizes who spoke when in a given segment of audio \cite{Bredin23}.  Meanwhile, intermediate embeddings, such as i-vectors and x-vectors, extracted from deep neural networks allow for recognizing speakers \cite{sujiya2017}.

While these tools provide state-of-the-art performance for their respective speech tasks, these tools have limited applicability to analyzing social interactions. Furthermore, non-verbal communication, such as facial expressions, body language, gestures, and posture, play a significant role in face-to-face communication and are analyses beyond what current speech processing tools can provide. Through the addition of inertial data, our work is different than existing speech processing works as we aim to investigate the dynamic and interactive processes involved in social interactions, namely gestures and body movements. Lastly, we do not employ any natural language processing on the input audio, which increases the privacy of the user's spoken content.

%Furthermore, these speech processing models are large and require significantly more computational resources than provided by a single commodity smartwatch. For instance, the base Whisper model has 74M parameters \cite{radford2022whisper}. Using float16 quantization, the model would still require 142MB of disk memory and around 500MB of RAM \cite{whispercpp}. While the primary smartwatches used in this work have up to 1GB of RAM for all their processes, the model would run slowly at just half-precision and present a cumbersome load on concurrent processes. As will be discussed later in section \ref{smartwatch-deployment}, our optimized model is only 857KB in size, which is over two orders of magnitude smaller than the size of Whisper, and has an inference time of less than 1s on smartwatches. Since the focus of our work is conversation detection using a single commodity smartwatch, incorporating large models like Whisper in the pipeline is not suitable given the computational needs of these speech recognition models.

%% file: conversation_modeling.tex
\label{conversation_modeling}
In this work, we aim to sense social interactions, specifically face-to-face conversations. Previous works in detecting face-to-face conversations have varied in their approach to modeling conversations as there is significant variability in real-world social interactions \cite{warren2006features}. Rahman \textit{et al.} \cite{rahman2011mconverse} defined conversation episodes to consist of user speaking and listening events while other works used conversational turn-taking as the fundamental unit of conversations \cite{lee2013sociophone, liang2023automated, hsiao2012recognizing}. Not all speech constitutes conversation; however, there is common agreement in literature that conversations across languages and cultures are characterized by turn-taking \cite{holler2016turn, donaldson1979one, sacks1975simple}. Thus, we also formulate this task around conversational turn-taking involving the device user. We approach the task as a three-class classification problem in line with existing literature and previous work \cite{liang2023automated}. The three classes are: 1) conversation, 2) other speech, and 3) background noise.

The \textit{conversation} class is defined to be instances where spoken communication with turn-changes occurs between the participant wearing the smartwatch and at least one other participant in the study. The \textit{other speech} class is defined as instances where spoken communication does not involve the participant wearing the smartwatch or foreground speech by the smartwatch user that does not contain turn-changes. Additionally, this class captures instances of when a participant wearing a smartwatch stops participating in a group conversation. Lastly, the \textit{background noise} class contains instances where there is no face-to-face spoken communication in the foreground. That is, this class captures speech from music, TV, or spoken communications in the background.

%% file: data_collection.tex
To develop and evaluate our approach, we collected a labeled dataset with synchronized audio and inertial data from a wrist-worn device during social activities in acoustically challenging environments. This multimodal dataset does not exist in literature, which prompted us to create one. This dataset can further open research avenues in understanding movement patterns during face-to-face communication, improving human-machine interaction by recognizing social cues, and more. This section presents the data collection process realized through two IRB-approved user studies - one performed in the \textit{laboratory} and one performed in \textit{semi-naturalistic} settings. In each study session, groups of two to four participants engaged in a set of social activities. We first present the hardware setup, followed by the data collection and annotation protocols.

\begin{comment}
\begin{table}
\caption{Study participant and group details (L: Left, R: Right, M: Male, F: Female, SW: Smartwatch).}
\begin{tabular}{c|c|c|ccc|ccc} 
  Group \#& Group Size&Setting&\multicolumn{3}{c}{SW User 1}& \multicolumn{3}{c}{SW User 2} \\ \hline
 
 & && \makecell{Handed-\\ness}&\makecell{Watch\\Hand}&Gender& \makecell{Handed-\\ness}&\makecell{Watch\\Hand}&Gender \\ 
 \hline
 1 & 3&Lab&R&L&   M& R& L&M \\ 
 2 & 3&Lab&R& L&F& R& L& M\\ 
 3 & 2&Lab&R& L&  M& R& L& M\\ 
 4 & 3&Lab&R& L&  M& R& L& M\\ 
 5 & 3&SN (lobby) & R& L&M& R& L& M\\ 
 6 & 3&SN (lobby) & R& L&M& R& L& M\\ 
 7 & 3&SN (lobby) & L& L&F& L& L& F\\ 
 8 & 4&SN (outdoors) & R& L& F& L& R& F\\ 
 9 & 3&SN (lobby) & R& R& M& R& L& F\\ 
 10 & 4&SN (lobby) & R& R& F& R& L& M\\ 
 11 & 4&SN (outdoors) & L& L& F& R& L& M\\ 
\label{table:group_details}
\end{tabular}
\vspace{-10pt}

\end{table}
\end{comment}

\begin{comment}
\begin{table}
\caption{Distribution of participants' handedness and watch wrist.}
\label{table:group_summmary}
\begin{tabular}{ c|c|c |c} 
%\multirow{2}{*}{}
\multicolumn{2}{ c }{}& \multicolumn{2}{ c }{Watch Wrist}\\ 
\cline{3-4}
 \multicolumn{2}{c}{}& Left&Right\\
\hline
\multirow{2}{6em}{Handedness} & Left-hand Dominant& 3&1\\ 
& Right-hand dominant& 16&2\\ 
\end{tabular}
\vspace{-10pt}
\end{table}
\end{comment}
\subsection{Hardware Setup}
We used one Fossil Gen 4 smartwatch and one Fossil Gen 5 smartwatch to collect data from participants. Both smartwatches are equipped with a Qualcomm Snapdragon 3100 processor and driven by Google's WearOS operating system. The smartwatches also have built-in accelerometer, gyroscope, and microphone sensors. On both watches, we collected audio and inertial data synchronously and saved data locally on the device using a custom-developed Android application. Lossless audio data was recorded at a sampling rate of 16kHz and 6-axis IMU (accelerometer and gyroscope) data was recorded at a sampling rate of 55Hz. We verified that differences in data collected between the two smartwatches were negligible. Post hoc, we downsampled the audio data into two additional sampling rates (1kHz and 2kHz) for model development and analysis, as further described in section \ref{data_preprocessing} and used in section \ref{downsampled}. We also recorded a video of each study session in its entirety for reference during the annotation process and facilitated the manual annotation process to be described in section \ref{data_annotation}.
\begin{table}
\centering
\caption{Study participant and group details (L: Left, R: Right, M: Male, F: Female, SW: Smartwatch).}
\label{table:group_details}

% Optimizations for height
\small                             % 1. Smaller font
\renewcommand{\arraystretch}{0.85} % 2. Tighter row spacing
\setlength{\tabcolsep}{4pt}        % 3. Tighter column spacing

\begin{tabular}{ccc |ccc|ccc} % Removed vertical lines for a cleaner, more compact look
\hline
\multirow{2}{*}{Group} & \multirow{2}{*}{Size} & \multirow{2}{*}{Setting} & \multicolumn{3}{c}{SW User 1} & \multicolumn{3}{c}{SW User 2} \\
 & & & Dom. Hand & Watch Hand & Gender & Dom. Hand & Watch Hand & Gender \\ 
\hline
1 & 3 & Lab & R & L & M & R & L & M \\ 
2 & 3 & Lab & R & L & F & R & L & M \\ 
3 & 2 & Lab & R & L & M & R & L & M \\ 
4 & 3 & Lab & R & L & M & R & L & M \\ 
5 & 3 & SN (lobby) & R & L & M & R & L & M \\ 
6 & 3 & SN (lobby) & R & L & M & R & L & M \\ 
7 & 3 & SN (lobby) & L & L & F & L & L & F \\ 
8 & 4 & SN (outdoors) & R & L & F & L & R & F \\ 
9 & 3 & SN (lobby) & R & R & M & R & L & F \\ 
10 & 4 & SN (lobby) & R & R & F & R & L & M \\ 
11 & 4 & SN (outdoors) & L & L & F & R & L & M \\ 
\hline
\end{tabular}
\vspace{-10pt}
\end{table}
\subsection{Data Collection Protocol}
\label{data_collection}
We collected data from 11 groups of participants across \textit{lab} and \textit{semi-naturalistic} settings for a total of 35 participants. Each group had a unique set of participants that did not overlap with any other groups. The \textit{lab} setting was a quiet, acoustically-controlled environment. The \textit{semi-naturalistic} settings included the lobby of a busy academic building in which there were background conversations and non-speech sounds from sources such as elevators and rolling utility carts, and an outdoors patio café in which there were background conversations and non-speech sounds from sources such as wildlife and engines.

To better quantify the acoustic characteristics of the data collection environments, we measured the A-weighted, equivalent continuous sound level (LAeq) of the environments without participant activity using the National Institute for Occupational Safety and Health (NIOSH) Sound Level Meter application on an iPhone, which is compliant with sound level meter standards \cite{CELESTINA2018119, niosh}. LAeq, measured in decibels, is the average sound energy over a period of time that emphasizes frequencies perceived by humans and is commonly used as a standard metric of noise levels. The \textit{lab} setting without participant activity had an average LAeq of 50.7dBA. The lobby of the \textit{semi-naturalistic} setting had an average LAeq of 70.2dBA and the outdoors patio café had an average LAeq of 60.3dBA. For reference, rainfall is around 50dBA, a normal conversation is around 60dBA, and a washing machine is around 70dBA \cite{noiselevels}. These sound levels show the acoustic variations of the data collection environments and specifically highlight the acoustically challenging nature of the \textit{semi-naturalistic} environments.

Each group consisted of two to four participants. Participants were diverse in gender and cultural representation, with participants of American, Chinese, Indian, and Australian backgrounds. This is important as communication styles, especially non-verbal communication, vary across cultures \cite{matsumoto2016cultural}. Details on the composition of the groups are in Table \ref{table:group_details}. %Some groups consisted of participants who knew each other while other groups consisted of participants who met for the first time. 

Within each group, two participants wore the data-collecting smartwatches. To increase the ecological validity of the study, the two participants were instructed to wear the smartwatch on whichever wrist they would normally wear a watch. All group participants clapped at the beginning of the recording process to synchronize audio and inertial data for data annotation and processing. Within their groups, a researcher asked all participants to perform the following activities:
\begin{enumerate}
    \item \textbf{Group Conversation: }All participants of the group played a NASA decision-making survival game while sitting \cite{hall1970effects}. In the \textit{lab} sessions, participants sat in chairs around a whiteboard, while in semi-naturalistic sessions, participants sat in chairs around a table. A participant not wearing the smartwatch was tasked with recording the group's responses on a whiteboard in the \textit{lab} sessions and on a sheet of paper in the \textit{semi-naturalistic} sessions. To emulate instances where individuals are situated in group conversations but not actively participating in the discussion, one participant wearing a smartwatch was instructed to discontinue speaking partway through the game and only listen.% while others continued discussing.
    \item \textbf{Group Conversation While Eating: } All participants were given snacks and instructed to chat with each other on any topic of their choosing while eating their snacks. Many social gatherings involve food and take place at restaurants or other venues that can be acoustically noisy. The purpose of collecting data with this activity was: 1) to simulate these louder environments in which social interactions occur and 2) to capture hand movements from eating in order to understand differences in hand movements due to speech-related gestures versus eating. Similar to activity 1, one participant wearing a smartwatch was instructed to discontinue speaking partway through the group conversation. 
    \item \textbf{Listening to Music: } Using a speaker, the researcher played music from a variety of musical genres. All participants listened to the music for 2-3 minutes.
    \item \textbf{Reading Out Loud:} Each participant wearing a smartwatch read out loud a random passage from one of three non-fiction books for two minutes. Then, a researcher played music different than the music played in activity 3, while the participant continued reading for another 2 minutes. 
    \item \textbf{Watching TV:} All participants watched two 3-5 minute video clips from a set of pre-selected clips from TV shows, talk shows, sportscasts, and documentaries, all of which contained conversations or narrations. Participants set the volume of the video clip playback, and conversation between participants was allowed. 

\end{enumerate}

\begin{figure}
\includegraphics[width=10cm]{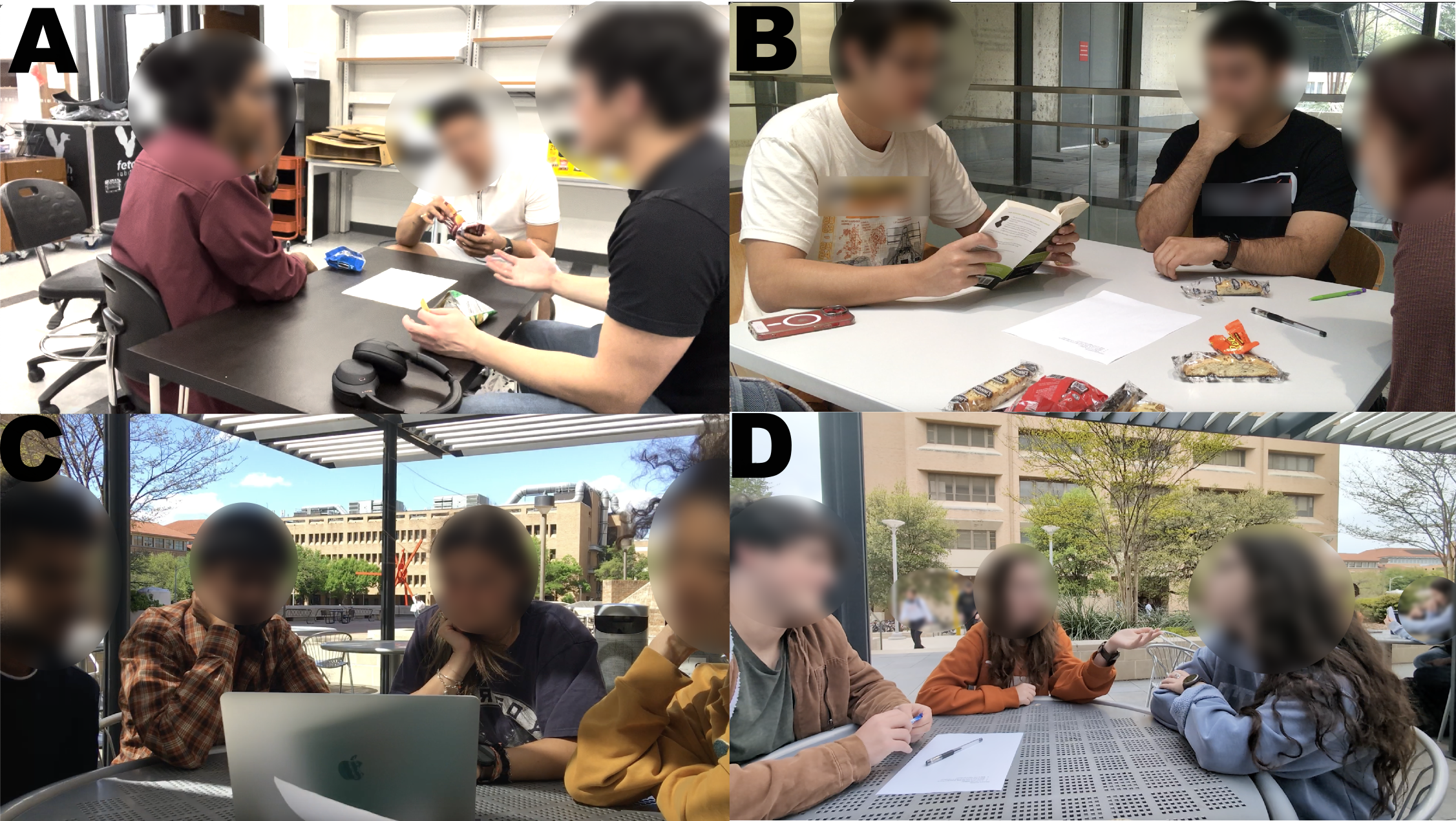}
\caption{Participants from different sessions performing group activities performed in the \textit{lab} and in a \textit{semi-naturalistic} setting. A: Participants having conversations while eating in a \textit{lab} setting. B: Participants reading out loud in a \textit{semi-naturalistic} setting. C: Participants watching a video in an outdoors \textit{semi-naturalistic} setting. D: Participants playing a team building exercise in a outdoors \textit{semi-naturalistic} setting. }
\label{study-fig}
\vspace{-10pt}

\end{figure}

This set of activities was chosen for being acoustically challenging yet representative of activities that take place in daily life. In total, we had a total of 35 participants across all study sessions and collected audio and inertial data from 22 participants. Our annotated dataset contains a total of 14.6 hours of audio and inertial data. Figure \ref{study-fig} shows screenshots of videos recorded during each study session, highlighting the setting and nature of activities. 

\subsection{Data Annotation Process}
\label{data_annotation}
After data collection, one researcher (one of the paper authors) manually annotated participants' audio and inertial data. We used ELAN \cite{wittenburg2006elan} to annotate audio while referencing the recorded video for ground truth and followed an annotation scheme established in prior works as discussed in section \ref{conversation_modeling}. We initially examined the social event detection problem at a granularity of 10-second segments. Consequently, we assigned a label, either \textit{conversation}, \textit{other speech}, or \textit{background noise}, to each 10-second segment of audio and inertial data. For 10-second segments that contained more than one class of activity, we assigned the segment with the class that occupies the majority of the 10-second segment. For instance, in a 10-second segment that contains 7 seconds of conversation and 3 seconds of noise, the 10 second segment is assigned the \textit{conversation} label. 

As will be discussed in detail in section \ref{overall-frame-sensitivity}, we found that our approach achieves optimal performance at window lengths of 30 seconds through a sensitivity analysis. Therefore, we aggregated our 10-second labels and applied them to 30-second segments as appropriate following the same method of assigning the 30-second segment with the label of the majority-duration class. The remainder of our paper discusses our framework with 30-second window lengths.

\begin{figure}
\includegraphics[width=10cm]{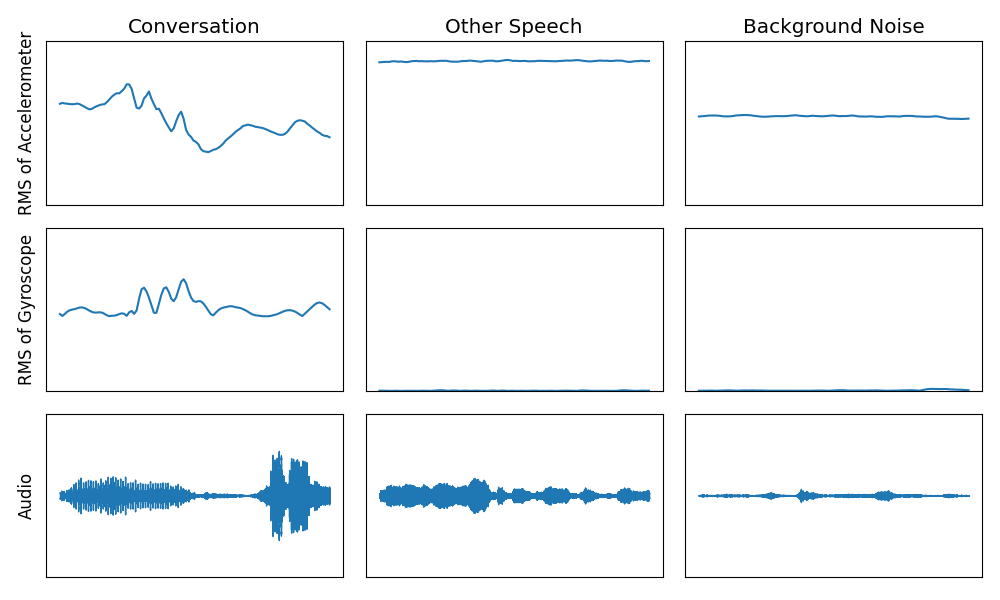}
\caption{Example raw acoustic and inertial data of one participant across all three classes.} %The inertial data reflects that the participant gestured while speaking in a conversation and stayed relatively motionless in the other two classes.}
\label{raw-data}
\vspace{-10pt}

\end{figure}

%% file: social_event_detection.tex
In this section, we present the data preprocessing and models developed for our conversation detection framework in acoustically challenging environments.

\subsection{Data Preprocessing}

\label{data_preprocessing}
\subsubsection{Audio preprocessing} As described in section \ref{data_collection}, we collected the input audio at a sampling rate of 16 kHz. To obtain an audio dataset at target sampling rates of 1khz and 2khz, we downsampled the original audio to each target sampling rate. In order to maintain a consistent shape of the audio data throughout the 16kHz, 2kHz, and 1kHz datasets for model development and evaluation, we interpolated the downsampled audio through high-quality FFT-based bandlimited interpolation to achieve a data shape of the downsampled audio identical to that of the original 16kHz audio. Despite having the same data shapes across the 16kHz, 2kHz and 1kHz datasets, speech intelligibility is significantly degraded in the 2kHz and 1kHz datasets. 

Within each 30-second raw audio segment, we calculated the fast Fourier Transform (FFT) using a window length of 500ms and stride of 250ms with 128 frequency bins inspired by prior works \cite{becker2019gestear, liang2023automated}. This yields image-like spectrogram features in a (128x120) shape per 30-second segment, regardless of the target audio sampling rate. We normalized these FFT features before feeding them into the model as inputs.
\label{audio_preprocess}

\subsubsection{IMU preprocessing} 
\label{imu_preprocessing}
The 6-axis IMU data was collected at a sampling rate of 55Hz. We standardized the values of each IMU axis to have a mean of 0 and standard deviation of 1. Within each 30-second segment, we framed the IMU data into frames of length 2 seconds with a 1 second overlap. This corresponds to 30 IMU frames of shape (6 x 110) within a 30-second segment of data. We then extracted statistical features from the raw IMU data and converted the raw IMU data into energy per channel.

\subsubsection{IMU Feature Selection}
\label{feature_selection}
For feature selection on the IMU data, we borrow the idea of mutual information from the field of information theory. The mutual information is a measure of dependency for two discrete random variables and is defined as: $$I(X;Y) = \Sigma_{x}\Sigma_{y}P(X, Y)log[\frac{P(X, Y)}{P(X)P(Y)}]$$ 
To calculate the mutual information between any given feature X and the target variable Y, where Y is the label for our three classes, we discretize the values of the feature. To achieve this, for every feature, we created a histogram with 10 bins and mapped each of the feature's values to its respective bin. Once the feature is discretized, we are able to calculate its mutual information, $I(X;Y)$, with the target variable. The higher the mutual information score between a feature and the target variable, the more the feature and target variable depend on each other. 

\begin{figure}
\includegraphics[width=10cm]{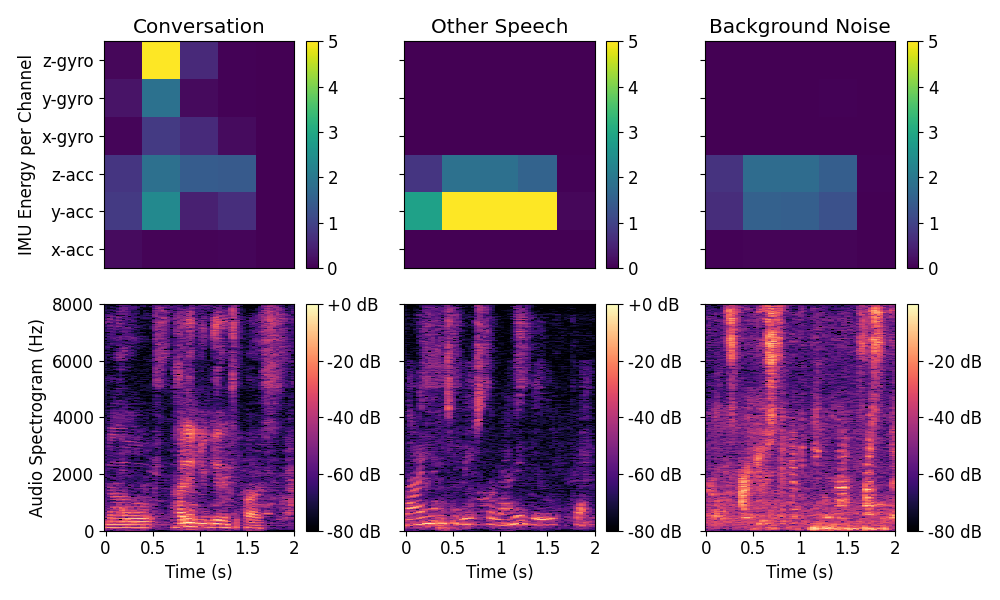}
\caption{Example inertial energies and acoustic spectrograms of one participant across all three classes. The audio and IMU data are synchronized.}
\label{spectrograms}
\vspace{-10pt}

\end{figure}

Since energy per IMU channel had a high mutual information score, we further explored using IMU energy distribution over time as a feature. We first transformed the normalized IMU signals into spectrograms by using the Short-time Fourier Transform (STFT). The STFT is a tool for transforming original time-domain signals to frequency-domain signals. The STFT of a time-series signal $x(t)$ is defined as: 
\[X(\tau,f) = \int_{-\infty}^{\infty}x(t)w(t-\tau)e^{-j2\pi ft} dt\]

where $w(t-\tau)$ is a windowing function. By taking the Fourier Transform of the original time-series signal $x(t)$ multiplied by the windowing function, we can localize in time the frequency content of the original signal. We calculated the STFT with 32 frequency bins at a time resolution of 400ms. We then calculated the energy per channel from the STFT (i.e. magnitude squared summed over all frequencies) to capture information on hand and wrist movements as they vary throughout social activities. This transformed the raw IMU signals into 30 image-like arrays of shape (6x5) per 30-second segment, as shown in Figure \ref{spectrograms}. 
%Lastly, we calculate the energy of the spectrogram by squaring the magnitude.

\subsection{Audio-only Models}
With data collected in section \ref{data_collection},  we explored audio models to establish primary results for face-to-face conversation detection using only audio inputs. Previous works have shown that convolutional neural networks (CNNs) when applied to FFT spectrograms of acoustic data are effective at detecting the presence of foreground speech \cite{nadarajan2019speaker}. Additionally, sequence models like long short-term memory networks have demonstrated capabilities in detecting speaker turns \cite{yin2017speaker}. 

\begin{figure}
\includegraphics[width=10cm]{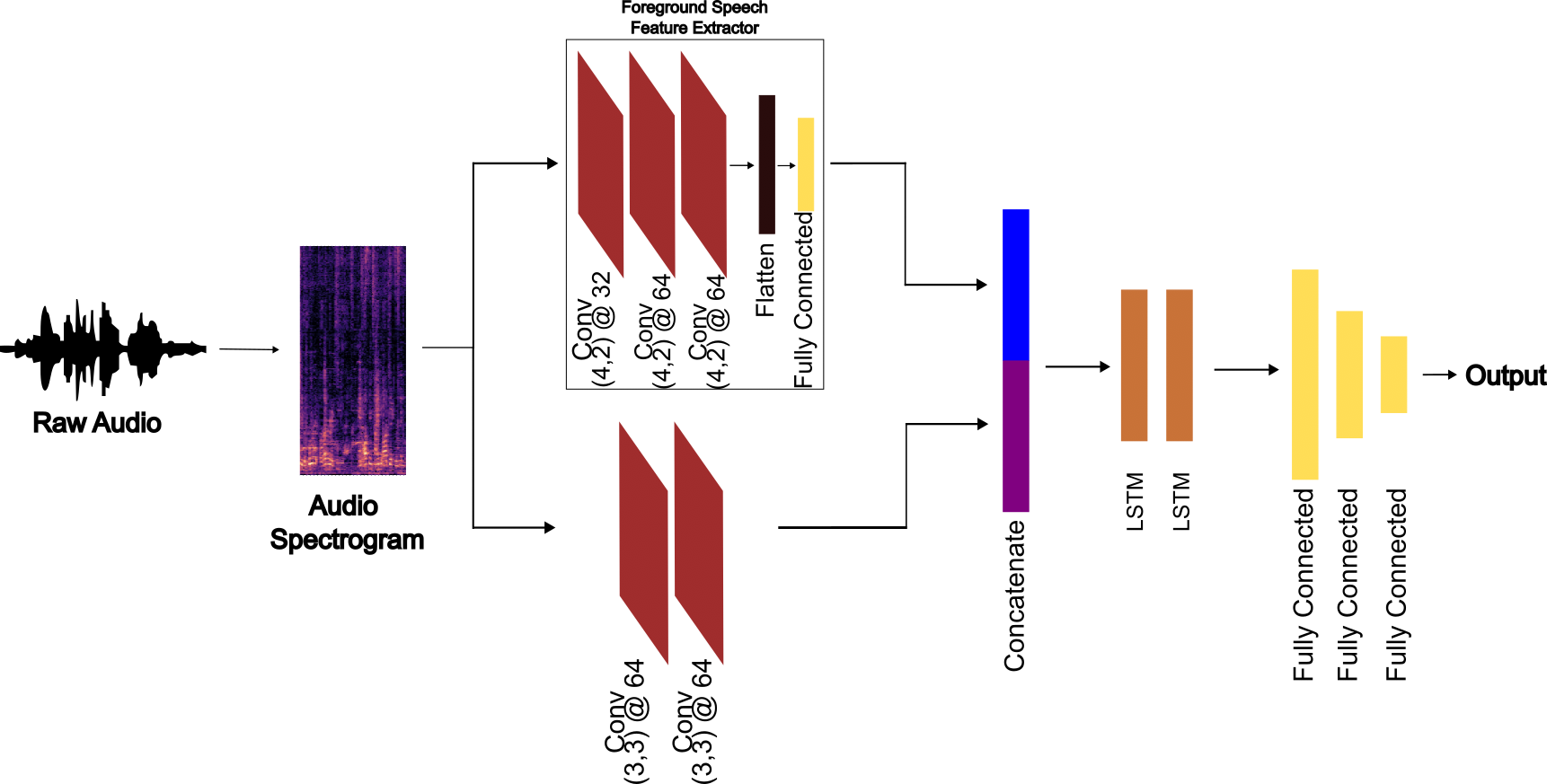}
\caption{Architecture of the audio classifier.}
\label{audio-only}
\vspace{-10pt}

\end{figure}

By common consensus on the definition of face-to-face conversations, the presence of both foreground speech and turn-taking is required \cite{sacks1975simple, donaldson1979one, holler2016turn}. Therefore, we built upon a state-of-the-art acoustic model that incorporates both the detection of foreground speech and speaker turns into a single architecture  \cite{liang2023automated}. In this acoustic model, the audio spectrograms are passed through a CNN that serves as a second feature extraction module by inferring the presence of foreground speech (Figure \ref{audio-only}). These foreground speech embeddings, along with embeddings extracted from the original audio spectrograms, are used as input features to a LSTM network to then capture the presence of foreground speaker turn changes. Three fully connected layers follow the LSTM network to make a final prediction on the input audio.

%\subsubsection{Pre-trained Models}
%Many sound classification models have been developed using Google's AudioSet dataset, a public dataset containing sound clips from over two million YouTube videos \cite{audioset}. The dataset contains 527 classes, including classes relevant to this work such as \textit{Conversation}, \textit{Chatter}, and \textit{Speech}. CNN14 is one such sound event detection model consisting of six convolution layers that has been trained and evaluated with AudioSet \cite{kong2020panns}. In this work, we leverage the pre-trained CNN14 to perform inference on our collected dataset.

%\subsubsection{Customized Models}
%Despite the state-of-the-art performance of acoustic models like CNN14 trained on AudioSet, there is a domain shift between audio extracted from YouTube videos and audio collected by a smartwatch in real life \cite{liang2019audio, liang2023automated}. Thus, we explore tailoring models specifically towards conversation detection using real-life user audio. Specifically, 

\subsection{Inertial-only Models}
Using motion data, we explored two neural network frameworks to establish an initial performance on our dataset.

\subsubsection{Neural Network Models} 
\begin{figure}
\includegraphics[width=10cm]{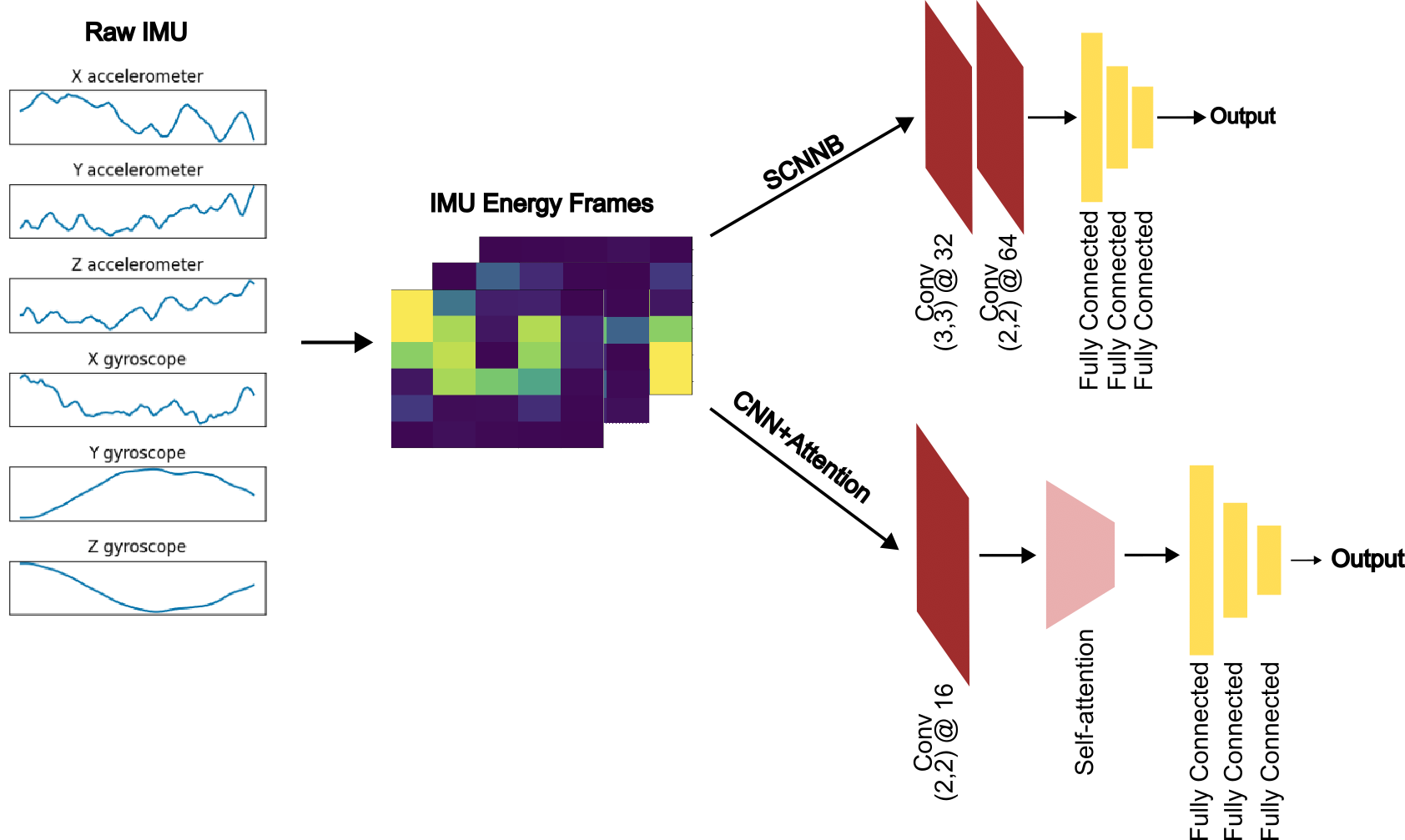}
\caption{Architectures of motion models explored in this work. Both models take as input IMU energy distributions over time, which has an image-like form.}
\label{imu_models}
\vspace{-10pt}

\end{figure}

The neural networks were inspired by and built upon the following models: (1) Shallow Convolutional Neural Network with Batch Normalization (SCNNB) \cite{lei2020shallow} and (2) Attend\&Discriminate \cite{abedin2021attend}. The architectures of both models are illustrated in Figure \ref{imu_models}. For both models, we empirically observed better performance with IMU energy as inputs to the model compared to using raw IMU data and therefore used IMU energy frames as model inputs. %Additionally, for each 30-second data segment, we pass 30 image-like IMU energy frames into the network. The network makes a classification probability on each IMU energy frame and the final layer averages the predictions across all 30 frames to return one final prediction on the 30-second data segment. 

\textit{SCNNB}: With the long-term vision of deploying a conversation detection model on edge devices with limited computational resources, we first experimented with utilizing a shallow, lightweight CNN. SCNNB is a network that achieves a performance on MNIST and CIFAR10 datasets comparable to deeper CNNs, such as MobileNets and VGGNet, with a shorter training time and fewer parameters. The network requires only a fraction of the time and space complexity required by larger CNNs and has motivated the use of shallow networks for HAR \cite{thakur2021feature}. Therefore, this model is suited for our task and eventual deployment onto edge devices. We leveraged SCNNB containing two convolution layers to extract features within the image-like IMU energy per channel that differentiate hand, wrist, and arm movements of the three target classes. 

%In our implementation of SCNNB, for each 30-second data segment, we pass 30 image-like IMU energy frames into the CNN. The CNN makes a classification probability on each IMU energy frame and the final layer averages the predictions across all 30 frames to return one final prediction on the 30-second data segment. 

\textit{Attend}\&\textit{Discriminate}: 
The second model we explored is Attend\&Discriminate, which has achieved state-of-the-art performance on public HAR datasets. The inputs are fed through a convolutional network to extract feature maps, which are then passed through a self-attention module to learn the interactions between sensor channels. The outputs are passed through a recurrent neural network to capture temporal information in the sensor channels and a temporal attention module to focus on the most relevant parts of the sequence, because time-steps do not always contribute uniformly to recognizing activities. 

We drew inspiration from the original Attend\&Discriminate model to guide our model development. The input to the model is the energy for each channel over time, and the network learns interactions between sensor channel energies. We modified the model architecture to remove the recurrent neural network and temporal attention module, as we observed significantly lower model accuracy with the inclusion of the sequence network and temporal attention module. We therefore call this model the \textit{CNN+Attention} network.
%Similar to our implementation of the \textit{SCNNB} and \textit{DeepConvLSTM} models, we found that model performance improved with IMU energy distributions over raw IMU data. 

\subsection{Fusion Methods and Models}
\begin{figure}
\includegraphics[width=10cm]{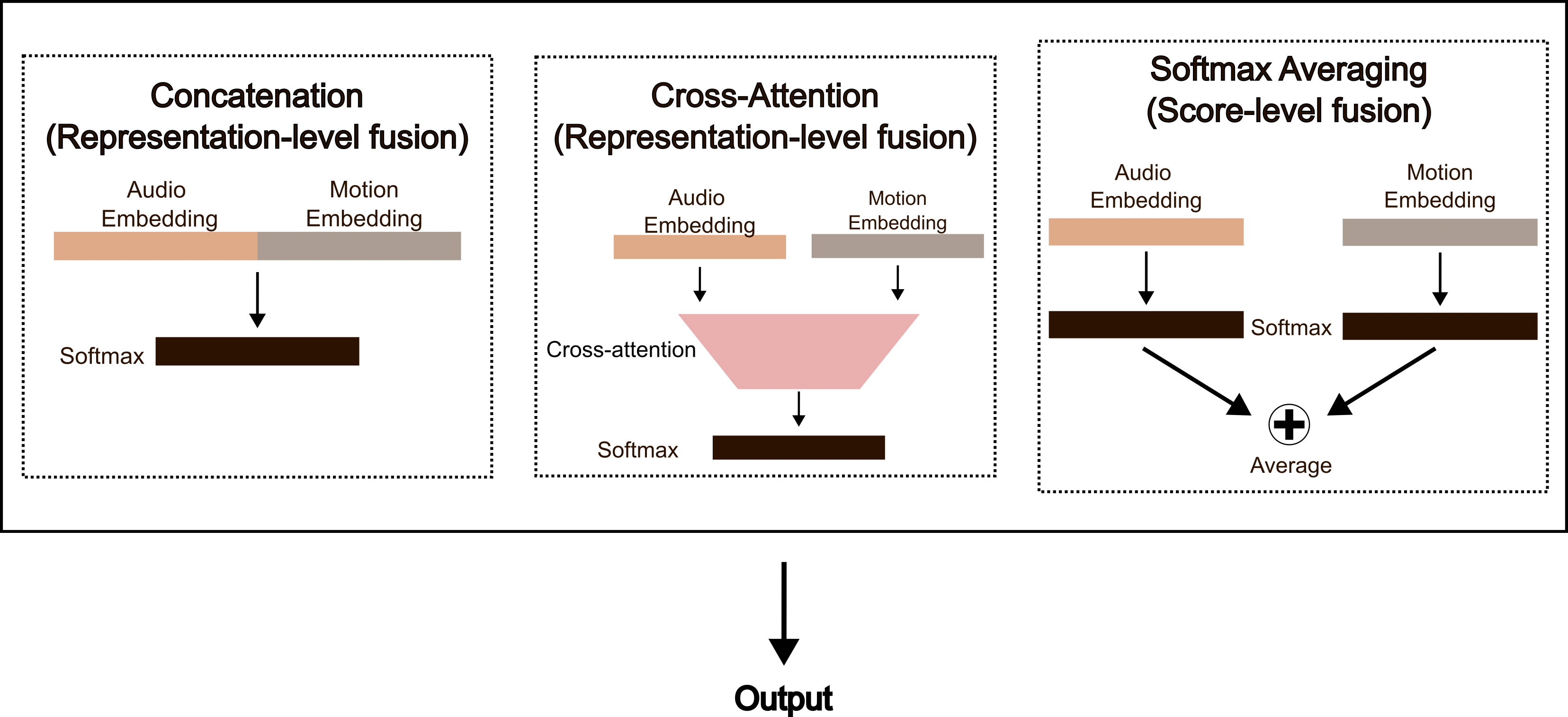}
\caption{An overview of fusion strategies explored in this work. Audio and inertial embeddings are first extracted from their respective networks. Representation-level fusion corresponds to combining embeddings of both modalities through concatenation or cross-attention. Score-level fusion through softmax averaging refers to training each modality's network separately and averaging the predicted probabilities.}
\label{fusion}
\vspace{-10pt}

\end{figure}

Standard techniques for fusing data from different sensor types include early-fusion and late-fusion \cite{pawlowski2023effective}. In early-fusion, data from each modality is concatenated together and the concatenated data is input into the machine learning model. In late-fusion, each modality is learned independently through separate networks and the learned representations are consolidated via an aggregation operation either at: 1) representation-level or 2) score-level. Representation-level fusion can be a concatenation of each modality's embeddings or a cross-modality attention module that captures the inter-modality relationships between each sensor's representations. This fusion is followed by a single classification head for joint training of each modality's network. In score-level fusion, each network is trained separately for each modality and the predicted class probabilities per network are averaged to obtain a final class prediction. The methods are illustrated in Figure \ref{fusion}.

In this work, since the acoustic and inertial data have different sampling rates and are preprocessed differently, simple concatenation of the raw audio and inertial data at the input stage is not possible. Thus, we shifted our focus to late-fusion. We experimented with different methods of fusing the acoustic and inertial embeddings and predicted class probabilities. Specifically, we fused the representations of the customized acoustic model with all six inertial models at different stages in order to better understand the impacts of data fusion for conversation detection.

\subsection{Model Optimization and Deployment}

%\subsubsection{Model Optimization}
To facilitate model deployment, we first optimized the model through quantization-aware training and weight pruning. With quantization-aware training, we lower the precision of model parameters from 32-bit float representations to 8-bit integer representations by introducing quantization effects during model training such that the trained model is more robust to the loss in weight precision. We additionally prune 50\% of the parameters per layer to remove insignificant parameters and obtain a sparser model. After optimization, we converted the model to TensorFlow Lite (TFLite), a data format that allows models to run on edge devices. Lastly, we developed an Android application using Java to load and invoke the model on the smartwatches. 
%Rather than training from scratch, we fine-tuned the pre-trained weights of the Pure-Acoustic Model with SCNNB model.

%\subsubsection{Deployment to Smartwatches}
\label{smartwatch-deployment}

%% file: eval_and_results.tex
Our objective is to analyze and evaluate the multimodal data obtained from a smartwatch for recognizing spoken, face-to-face conversations. By leveraging the dataset collected in section \ref{data_collection}, we seek to investigate the following questions:
\begin{itemize}
    \item What degree of information does each modality provide towards conversation detection?
    \item To what extent does the fusion of acoustic and inertial data contribute to conversation detection?
    \item How does a model pre-trained on data collected in a controlled \textit{lab} setting perform on unseen data collected in \textit{semi-naturalistic} settings?
\end{itemize}

\subsection{Evaluation Setup}
\label{eval-setup}
Conducting the study across both acoustically controlled and noisy environments and allowing participants to choose to wear the smartwatch on the wrist of their choice introduced a significant amount of variability in the collected data. Thus, we performed three sets of evaluations to better understand the role of multimodal data for conversation detection.  The three sets of evaluation are as follows: (1) evaluation in the \textit{lab-only} setting, (2) evaluation in the \textit{lab and semi-naturalistic} settings, and (3) evaluation in only \textit{semi-naturalistic} settings on \textit{lab}-trained models. Across all evaluation setups, we report the macro (unweighted) F1-score averaged across all groups used in each evaluation setup. The macro F1-score treats all target classes with equal importance, thereby removing the effects of imbalanced class distribution in the evaluation set. We obtain the 95\% confidence intervals for the F1-scores by bootstrap sampling the test dataset 200 rounds. 

In the \textit{lab-only} and \textit{lab and semi-naturalistic} evaluations, we explore how well the model generalizes across controlled and noisy settings, respectively. We followed a leave-one-group-out (LOGO) cross-validation scheme in which all but one group of participants were used for training and the remaining group was used for testing. This was repeated 4 times through all combinations in the \textit{lab-only} dataset and 11 times through all combinations in the \textit{lab and semi-naturalistic} combined datasets due to having 4 and 7 groups in the \textit{lab} and \textit{semi-naturalistic} settings respectively. To evaluate how well the \textit{lab} dataset generalizes to \textit{semi-naturalistic} settings, in the \textit{semi-naturalistic}-only evaluation we trained models using only \textit{lab}-collected data and evaluated models using only data collected from the \textit{semi-naturalistic} setting. Table \ref{results-summary} shows a summary of results obtained for each model across all evaluation setups.  
%reporting results for 2s windows with 1s overlap!!!

\subsection{Evaluation Results}

To assess the advantages of fusing acoustic and inertial data, we first  evaluated the performance of both the acoustic and inertial models separately (Table \ref{results-summary} and Table \ref{other-metrics}). For acoustic-based classification, the Pure-Acoustic Model  \cite{liang2023automated} achieved a macro-F1 score of 74.4$\pm$3.3\% on the \textit{lab} dataset. Among the motion models, the SCNNB model performs the best, reaching an F1 score of 53.7$\pm$2.4\% on the \textit{lab and semi-naturalistic} evaluation. 

In combining acoustic and inertial data, we observe an increase in performance across all models that employ concatenation for representation-level fusion of the acoustic and inertial embeddings. However, score-level fusion and representation-level fusion through attention did not improve upon the top single-modality classifier. Fusing the embeddings extracted from the Pure-Acoustic Model for the audio data and from the CNN+Attention architecture for the inertial data achieves the best F1-score of 82.0$\pm$3.0\% in evaluation in the \textit{lab} setting, representing a 7.6\%-point improvement in F1-score over the best single-modality classifier under the same evaluation setting. In \textit{lab and semi-naturalistic} evaluation, the fused audio model and SCNNB inertial model achieves the highest F1-score of 78.0$\pm$1.7\% yielding a 5.0\%-point increase in F1-score compared to the audio-only classifier. For the \textit{semi-naturalistic}-only evaluation, the fused audio and CNN+Attention inertial model again performs the best with an F1-score of 68.1$\pm$2.7\%, which is 3.2\%-points higher than that of the audio-only model. Furthermore, the decrease in F1-score in this evaluation setup shows the limitation of models trained entirely with data from acoustically controlled environments when evaluated on \textit{semi-naturalistic} contexts. Since the fused audio and CNN+Attention inertial model outperforms the fused audio and SCNNB model in two of the three evaluations, we consider the Pure-Acoustic Model with CNN+Attention through Concatenation to be the best performing multimodal classifier. %To reiterate, in the multimodal model, the audio data only passes through the Pure-Acoustic architecture as illustrated in Figure \ref{audio-only} and the IMU data only passes through the CNN+Attention architecture as illustrated in the bottom branch of Figure \ref{imu_models} before each modality's embeddings are brought together via fusion. Thus, in comparison to the audio-only and inertial-only models, we can analyze the impact of audio and IMU data in the multimodal model in Table \ref{results-summary}.

This improvement in multimodal model performance comes with only a 0.4\% increase in number of model parameters compared to the best audio-only classifier. The classifiers have 2.8K, 763.2K, and 766.5K parameters for the inertial-only (CNN+Attention), audio-only, and multimodal classifiers, respectively. This highlights the lightweight manner in which gestures and body movements captured by inertial data can be effectively incorporated for conversation detection.

\begin{table*}[htbp]
\small
\centering
\caption{Average macro-F1 score for each audio-only, motion-only, and audio plus motion model with all combinations of fusion strategies evaluated across three evaluation setups on the collected dataset. L-LOGO: Training on all but one lab session and evaluating on the holdout lab session. L+SN-LOGO: training on all but one session across the lab and semi-naturalistic sessions and evaluating on the holdout session. SN: training on all lab sessions and evaluating on all semi-naturalistic sessions.}

\label{results-summary}

\begin{tabular}{c c c c c c} 
\toprule
& Model& Fusion & L-LOGO& L+SN-LOGO&SN\\
\midrule
\multirow{1}{*}{Audio}
%& CNN14 & -& 22.0 & 23.2 & 23.1\\ 
& Pure-Acoustic Model \cite{liang2023automated}  & -& 74.4 $\pm$ 3.3 & 73.0 $\pm$ 2.0 & 64.9 $\pm$ 2.8\\ 
\hline
\multirow{2}{*}{Motion}
% & Random Forest& -& 35.6 & 38.4 & 33.0\\
% & Gaussian Naive Bayes& -& 39.9 & 36.0 & 43.3 \\
% & AdaBoost& -& 39.7 & 47.4 & 45.6\\
 & SCNNB  & -& 51.6 $\pm$ 4.1 & 53.7 $\pm$ 2.4 & 49.0 $\pm$ 3.1\\
% & DeepConvLSTM & -& 35.0 & 30.7 & 35.0\\
 & CNN+Attention  & -& 49.6 $\pm$ 4.3 & 50.8 $\pm$ 2.5 & 48.7 $\pm$ 3.2\\
\hline
 \multirow{6}{*}{Audio+Motion}
 %& Pure-Acoustic Model  + Random Forest& Softmax Averaging& 42.8 & 31.9 & 47.1\\
 %& Pure-Acoustic Model  + Gaussian Naive Bayes& Softmax Averaging& 43.4 & 40.9 & 45.1 \\
 %& Pure-Acoustic Model  + AdaBoost& Softmax Averaging& 56.3 & 49.1& 18.8\\
 & \multirow{3}{*}{\makecell{Pure-Acoustic Model  \\ + SCNNB}} & Softmax Averaging & 67.9 $\pm$ 2.2 & 61.4 $\pm$ 2.3 & 55.2 $\pm$ 2.7 \\
 % Softmax Averaging & 60.4 & 52.6 & 51.2\\
 & & Concatenation & 77.5 $\pm$ 3.0 & \textbf{78.0 $\pm$ 1.7} & 67.5 $\pm$ 2.5\\
 %& & Self-Attention & 52.7 & 45.9 & 65.9\\
 %& & Self-Attention & 62.2 $\pm$ 4.2 & 57.1 $\pm$ 2.3 & 66.1 $\pm$ 2.7\\
& & Self-Attention & 62.9 $\pm$ 3.8 & 57.9 $\pm$ 2.3 & 62.8 $\pm$ 2.9\\

 % \cline{2-6}
 %& \multirow{3}{*}{Pure-Acoustic Model  + DeepConvLSTM} & Softmax Averaging & 60.4 & 52.6 & 51.2\\
 %& & Concatenation & 75.1 & 77.1 & 66.3\\
 %& & Self-Attention & 52.1 & 48.7 & 47.1\\
 \cline{2-6}
 & \multirow{3}{*}{\makecell{Pure-Acoustic Model  \\ CNN+Attention}} & Softmax Averaging & 62.4 $\pm$ 3.6 & 65.2 $\pm$ 2.6 & 50.7  $\pm$ 2.6 \\
 %Softmax Averaging & 60.0 &  50.9 & 53.2 \\
 & & Concatenation & \textbf{82.0 $\pm$ 3.0} & 77.2 $\pm$ 1.8 & \textbf{68.1 $\pm$ 2.8}\\
 %& & Self-Attention & 63.0 & 60.3 & 58.5\\
 %& & Self-Attention & 66.0 $\pm$ 2.1 & 60.3 $\pm$ 2.3 & 56.3 $\pm$ 2.8\\
 & & Self-Attention & 59.6 $\pm$ 4.0 & 55.6 $\pm$ 2.5 & 55.8 $\pm$ 2.9\\

\bottomrule
\end{tabular}
\vspace{-10pt}

\end{table*}

\begin{table}
\centering
\footnotesize
\caption{Macro precision and recall. P: precision. R: recall. }
\label{other-metrics}

  \begin{tabular}{c c c c c c c c}
    \toprule
    {Modality}&
    {Model}&
      \multicolumn{2}{c}{L-LOGO} &
      \multicolumn{2}{c}{L+SN-LOGO} &
      \multicolumn{2}{c}{SN} \\
    \midrule
    & & P & R & P & R & P & R \\ \cline{3-8}
    Audio &Pure-Acoustic Model & 77.6 $\pm$ 3.4 & 74.2 $\pm$ 3.0 & 73.5 $\pm$ 2.1 & 73.4 $\pm$ 1.9 & 65.3 $\pm$ 2.4 & 64.8 $\pm$ 2.8 \\ \hline
    \multirow{2}*{IMU} &SCNNB & 55.8 $\pm$ 5.0 & 51.8 $\pm$ 3.5 & 58.9 $\pm$ 2.7 & 53.5 $\pm$ 2.2 & 50.2 $\pm$ 3.4 & 48.8 $\pm$ 3.0 \\
     &CNN+Attention & 56.0 $\pm$ 5.4 & 49.8 $\pm$ 3.5 & 59.7 $\pm$ 3.0 & 51.3$\pm$ 2.1 & 48.9 $\pm$ 3.0 & 48.9 $\pm$ 3.1 \\ \hline
    \multirow{3}*{Audio+IMU} & \makecell{Pure-Acoustic Model \\ + SCNNB }& 79.4 $\pm$ 3.6 & 77.3 $\pm$ 3.0 & \textbf{78.2 $\pm$ 1.8} & \textbf{78.3 $\pm$ 1.7} & 69.9 $\pm$ 2.6  & 67.6 $\pm$ 2.6 \\
     &\makecell{Pure-Acoustic Model \\ + CNN+Attention} & \textbf{82.4 $\pm$ 3.1} & \textbf{82.7 $\pm$ 2.7} & 77.7 $\pm$ 2.0 & 77.0 $\pm$ 1.8 & \textbf{71.1 $\pm$ 2.6} & \textbf{68.2 $\pm$ 2.5} \\
    \bottomrule
  \end{tabular}
  \vspace{-10pt}

\end{table}

\begin{figure}
\includegraphics[width=\linewidth]{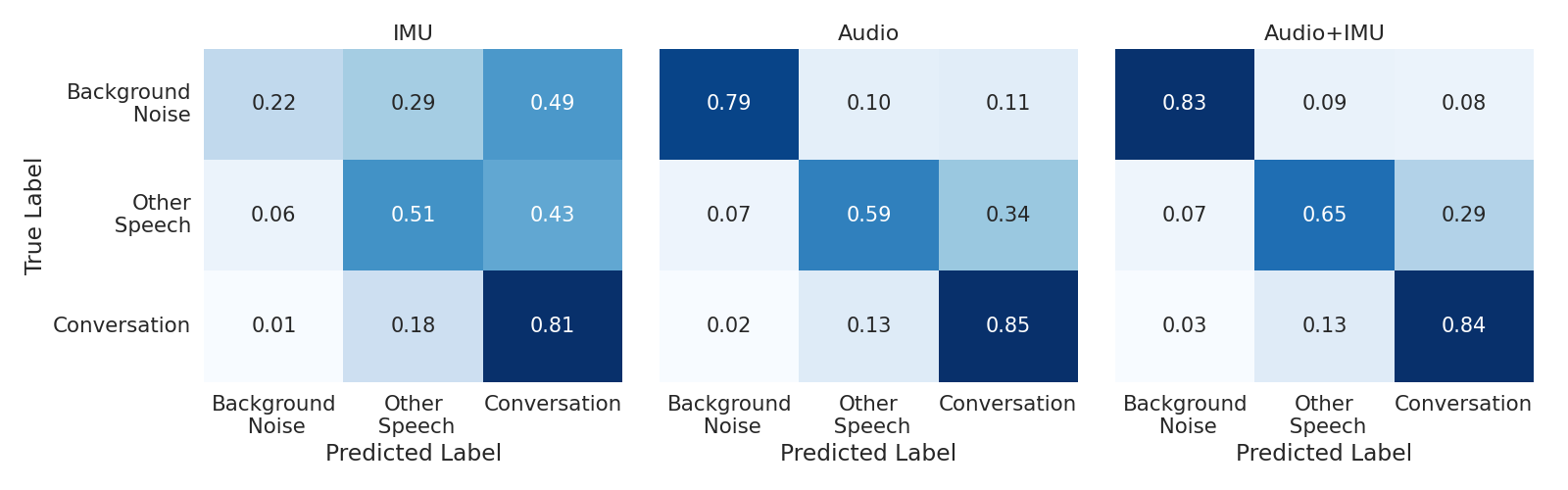}
\caption{Confusion matrices for inertial-only (left), audio-only (center), and audio-inertial frameworks (right).}
\label{confusion_matrices}
\vspace{-10pt}

\end{figure}

\begin{figure}
\includegraphics[width=0.9\linewidth]{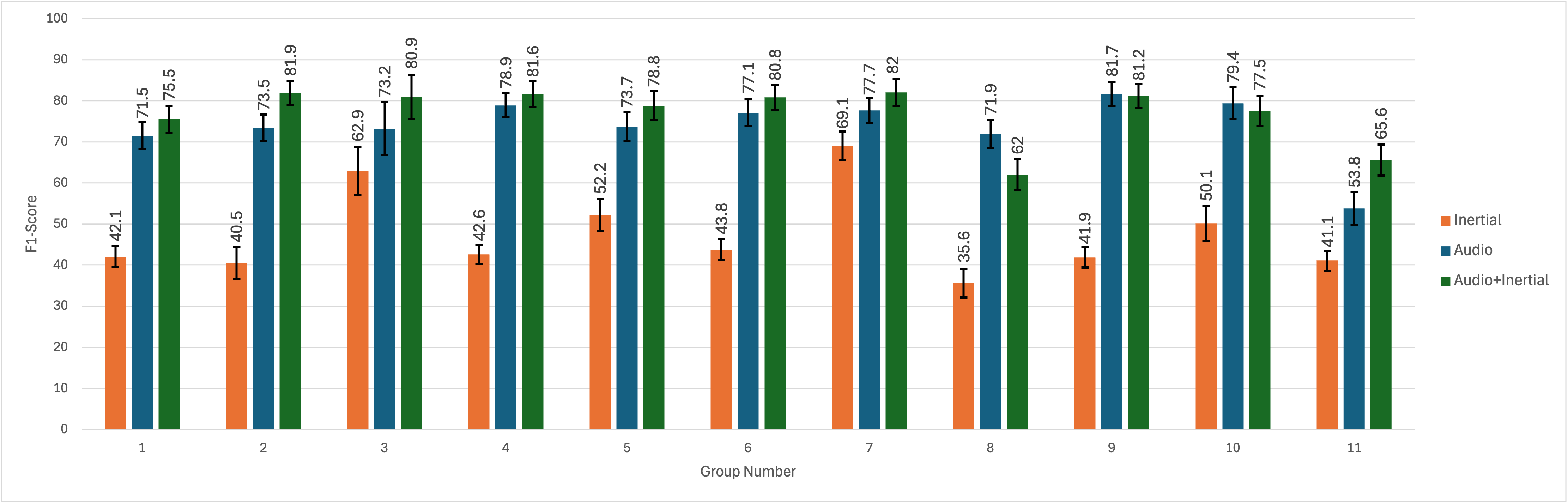}
\caption{A comparison of LOGO evaluation results (macro F1-score) across audio, inertial, and multimodal classifiers. The multimodal framework improves upon any single-modality classifier in all but three groups.}
\label{perGroup}
\vspace{-10pt}
\end{figure}

Confusion matrices comparing the performance of the single modality classifiers composing the top multimodal classifier and the multimodal classifier itself are shown in Figure \ref{confusion_matrices}. Per-group performance for the audio (Pure-Acoustic Model), inertial (CNN+Attention), and multimodal (Pure-Acoustic Model with CNN+Attention through Concatenation) models are shown in Figure \ref{perGroup}. The \textit{lab} groups have an average F1-score of 80.0$\pm$3.7\%  while the \textit{semi-naturalistic} groups have an average F1-score of 75.4$\pm$3.4\% , though group 7 with the highest F1-score of 82.0$\pm$3.2\%  comes from the \textit{semi-naturalistic} setting. On the other hand, groups 8 and 11 have the worst F1-scores of 62.0$\pm$3.8\% and 65.6$\pm$3.8\% respectively. Upon examination of these two groups, we found that the declined performance could be due to two factors. First, both smartwatch users in group 8 and one smartwatch user in group 11 wore the watch on their dominant hand, unlike most other participants. Second, groups 8 and 11 were the only groups whose data collection was outdoors. Acoustic and motion artifacts unique to the outdoors setting in their small sample size could have degraded model performance. 


\subsection{Real-Time Smartwatch Implementation}
%As previously mentioned, we used one Fossil Gen 4 smartwatch and one Fossil Gen 5 smartwatch as data collection devices. The Gen 4 smartwatch has 512MB RAM and 4GB storage while the Gen 5 smartwatch has 1GB RAM and 8GB storage. 

Compared to smartphones or other edge devices like Raspberry Pis, smartwatches have significantly fewer computational resources. Despite these computational limitations of smartwatches, we demonstrate that our conversation detection model can deploy to smartwatches.

We focus our model deployment discussion on the Pure-Acoustic Model with SCNNB architecture after discovering hardware limitations in the smartwatches to support the Pure-Acoustic Model with CNN+Attention model architecture. Specifically, we discovered that some operations required for the attention mechanism are not supported by the hardware in the Fossil Gen 4 and 5 smartwatches. Additionally, while the audio-inertial model can run on both the Fossil Gen 4 and 5 smartwatches, we focus our deployment evaluation and analysis on the Gen 5 smartwatch as it is newer than the Gen 4 smartwatch (2019 vs 2018) and has double the RAM (1GB vs 512MB). 

%For comparison to the Pure-Acoustic Model, we also optimize the audio-only model in the same manner. 
We evaluate the optimized Pure-Acoustic Model with SCNNB performance in the \textit{lab and semi-naturalistic} LOGO evaluation setups and show the results in Table \ref{tab:app_inference}. For comparison to the Pure-Acoustic Model, we also optimize and evaluate the audio-only model. We observe that the joint audio-inertial optimized model achieves a performance similar to its pre-optimization performance and still outperforms the optimized, audio-only model.

We also profile the inference time of both the audio-only and joint audio-inertial models to understand the potential real-time applications of our system. We summarize the TFLite size and average inference times of both models while running on the Fossil Gen 5 smartwatch in Table \ref{tab:app_inference}. The average inference time is measured across 10 successive invocations of the model on the smartwatch. 

As many smartwatches have been released since the Fossil Gen 5 smartwatch in 2019, we also profiled the model's inference time on a newer smartwatch, the Google Pixel Watch 2 released in 2023. With 2GB RAM, the average inference time of the joint audio-inertial model is reduced by over a factor of two down to 400ms. Furthermore, this smartwatch hardware supports the Pure-Acoustic Model with CNN+Attention model architecture with a similar runtime to the Pure-Acoustic Model with SCNNB model. %These explorations indicate that our joint acoustic-inertial sensing framework can run on smartwatch hardware from over five years ago, and as hardware support improves, the runtime of the deployed model will improve as well. 

\begin{comment}
\begin{table}
    \centering
    \caption{Comparison of the performance, size, and inference time on a Fossil Gen 5 smartwatch of the optimized audio-only and audio-inertial models.}

    %\begin{tabular}{clcc}
    \begin{tabular}{@{}c *{3}{C{2cm}} @{}}
    \toprule
         Model&   F1-score &TFLite Size (kB) & Inference Time on Smartwatch (ms)\\
         \hline
         Optimized Pure-Acoustic Model&  74.6 $\pm$ 2.0 &831 & 953.6 \\
         Optimized Pure-Acoustic Model + SCNNB &  77.3 $\pm$ 1.8 &857 & 972.5 \\
    \bottomrule
    \end{tabular}
    \label{tab:app_inference}
    %\vspace{-10pt}
\end{table}
\end{comment}
\begin{comment}
\begin{table}[t]
    \centering
    \caption{Performance, model size, and inference latency on a Fossil Gen~5 smartwatch for optimized audio-only models.}
    \label{tab:app_inference}

    \setlength{\tabcolsep}{8pt}
    \renewcommand{\arraystretch}{1.2}

    \begin{tabular}{@{}cccc@{}}
        \toprule
        \textbf{Model} 
        & \textbf{F1-score (\%)} 
        & \textbf{TFLite Size (kB)} 
        & \textbf{Inference Time (ms)} \\
        \midrule
        Optimized Pure-Acoustic 
        & $74.6 \pm 2.0$ 
        & 831 
        & 953.6 \\
        Optimized Pure-Acoustic + SCNNB 
        & $77.3 \pm 1.8$ 
        & 857 
        & 972.5 \\
        \bottomrule
    \end{tabular}
    \vspace{-10pt}
\end{table}
\end{comment}

\begin{table}[t]
    \centering
    \caption{Performance, model size, and inference latency on a Fossil Gen~5 smartwatch for optimized models.}
    \label{tab:app_inference}

    \small % Reduces font size and baseline skip
    \setlength{\tabcolsep}{5pt} % Slightly tighter column spacing

    \begin{tabular}{@{}lccc@{}} % 'l' is usually better for text labels than 'c'
        \toprule
        \textbf{Model} 
        & \textbf{F1 (\%)} 
        & \textbf{Size (kB)} 
        & \textbf{Latency (ms)} \\
        \midrule
        Pure-Acoustic 
        & $74.6 \pm 2.0$ 
        & 831 
        & 953.6 \\
        Pure-Acoustic + SCNNB 
        & $77.3 \pm 1.8$ 
        & 857 
        & 972.5 \\
        \bottomrule
    \end{tabular}
    \vspace{-10pt}
\end{table}

%\begin{figure}\centering
%\subfloat
%{\label{a}\includegraphics[width=.330\linewidth]{figures/audio_cm.png}}
%\subfloat
%{\label{b}\includegraphics[width=.330\linewidth]{figures/imu_cm.png}} 
%\subfloat
%{\label{c}\includegraphics[width=.330\linewidth]{figures/audio_imu_cm.png}}
%\caption{Confusion matrices for audio-only (left), inertial-only (center) and audio-inertial frameworks (right).}
%\label{confusion_matrices}
%\end{figure}

%% file: discussion.tex
In this section, we discuss additional evaluations performed across window lengths, activity contexts, audio sampling rates, and dataset environments and the tradeoff between ecological validity and participant handedness during the data collection study.

\subsection{Frame Sensitivity Analysis}
Frame sizes in HAR impact classification granularity, feature extraction, and model performance. Therefore, we gauged the impact of overall window length and IMU frame size for our novel approach on social interaction analysis in busy environments.

\subsubsection{Overall Frame Sensitivity}
\label{overall-frame-sensitivity}
\begin{figure}
    \centering
    \includegraphics[width=1\linewidth]{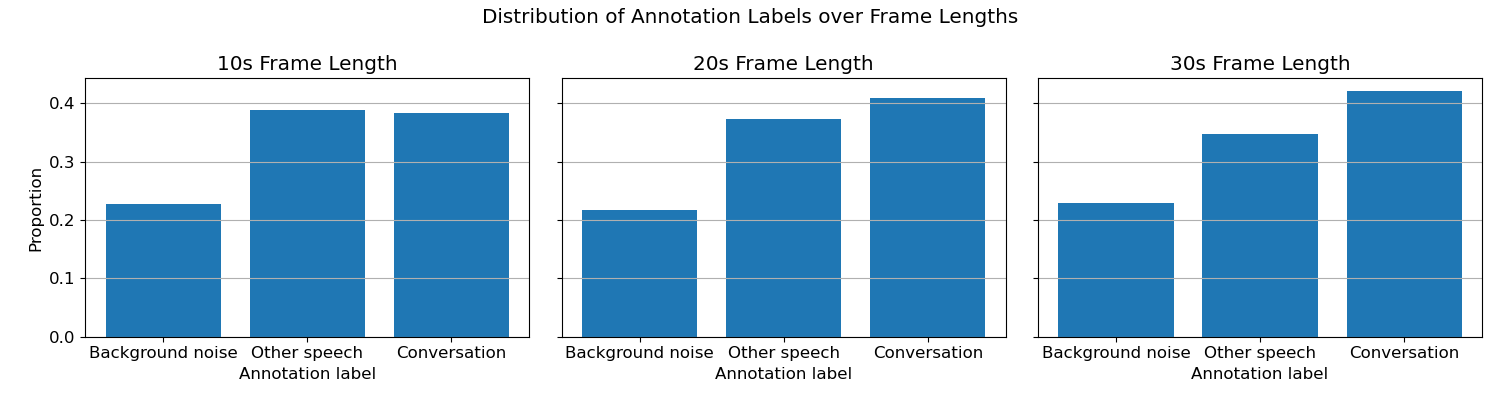}
    \caption{Distribution of label annotations across frame lengths.}
    \label{fig:label_distribution}
    \vspace{-10pt}
\end{figure}

Given the dynamic fluctuations characteristic of busy settings, we investigated the model's performance across window lengths in 10-second intervals spanning from 10 to 30 seconds. To contextualize the trade-off between classification accuracy and granularity, we first examined the distribution of social events across frame lengths in our dataset. Figure \ref{fig:label_distribution} shows that as the frame size increases, instances of \textit{other speech} become encapsulated by the \textit{conversation} label. This illustrates that larger window sizes can hide the fine-grained dynamics of an interaction, such as moments with a dominating speaker or lacking turn-taking, that give rise to the \textit{other speech} label in a 10-second window, but are incorporated into the \textit{conversation} label when the window size increases to 20 or 30 seconds due to the majority-voting labeling scheme. 

\begin{figure}[!htb]
   \begin{minipage}{0.59\textwidth}
     \centering
     \includegraphics[width=1\linewidth]{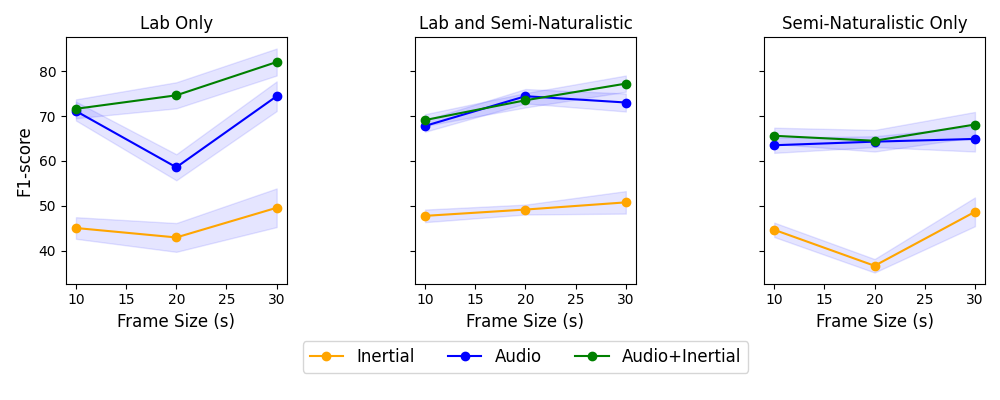}
     \caption{A comparison of macro F1-scores across all three evaluation setups for acoustic, inertial and multimodal classifiers with window lengths varying from 10 to 30 seconds.}
     \label{Fig:overall-frame-sensitivity}
   \end{minipage}\hfill
   \begin{minipage}{0.39\textwidth}
     \centering
     \includegraphics[width=1\linewidth]{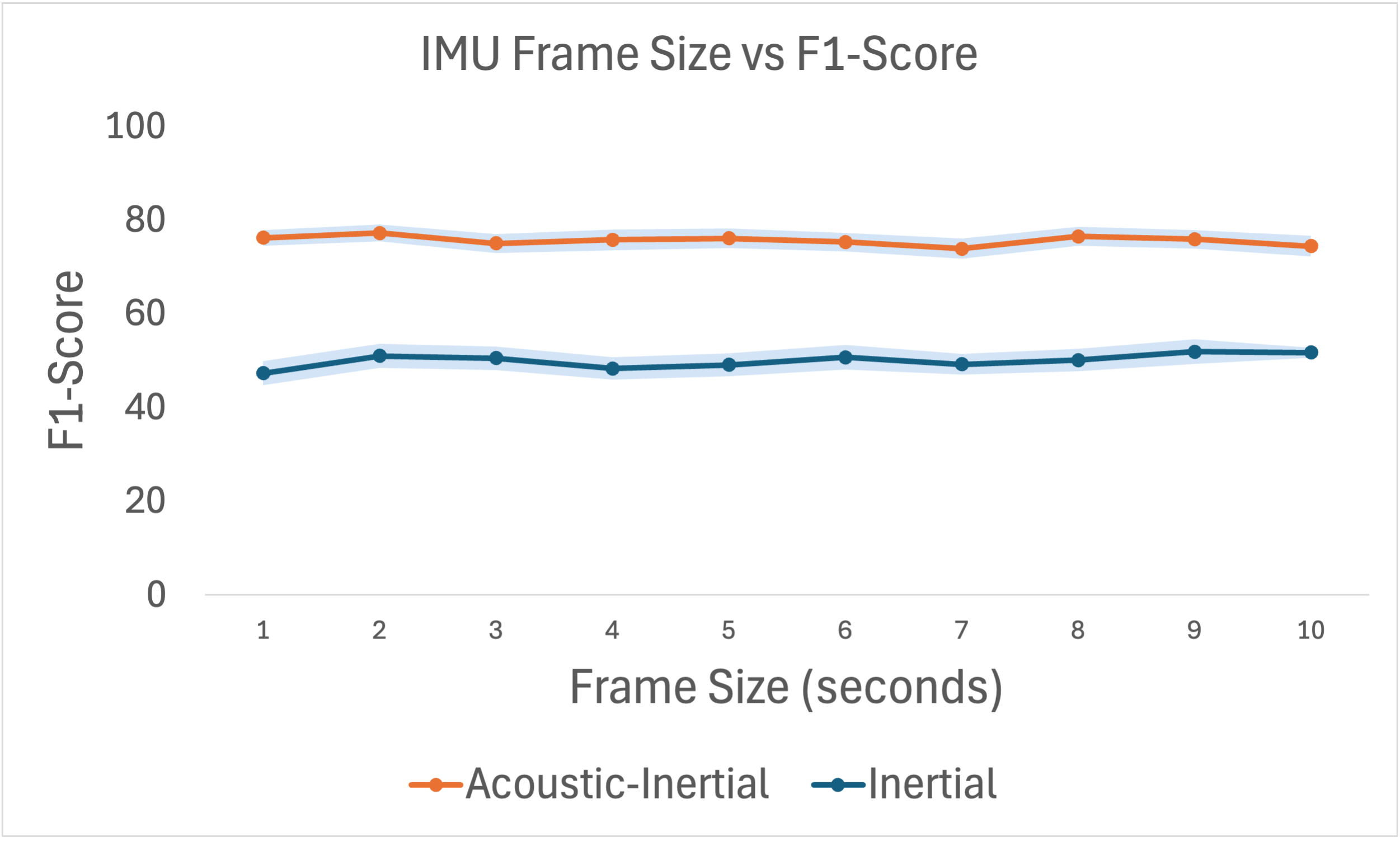}
     \caption{A comparison of \textit{lab+semi-naturalistic} LOGO evaluation results across inertial and multimodal classifiers with varying IMU frame sizes.}
     \label{Fig:imu-frame-sensitivity}
   \end{minipage}
   \vspace{-10pt}

\end{figure}

Like before, we evaluate the model with a LOGO cross-validation on \textit{lab-only} and \textit{lab and semi-naturalistic} data and with an evaluation on \textit{semi-naturalistic} data upon \textit{lab-only} training.  Figure \ref{Fig:imu-frame-sensitivity} shows that multimodal performance improves as window length increases, with peak performance at a window length of 30 seconds. This is in line with previous works that have found a 30 second window to provide the best tradeoff between classification accuracy and robustness \cite{rahman2011mconverse, bari2020automated, choudhury2003sensing, liang2023automated}. Although longer windows give better model performance, it reduces the granularity of conversation detection regarding the engagement dynamics of the conversation. %For instance, short conversations that give rise to the \textit{conversation} label in a 10-second window may become encapsulated by the \textit{background noise} label when the window size increases to 20 or 30 seconds. We specifically observed this occurrence when participants watched videos, as participants intermittently exchanged short remarks with each other regarding the video. Therefore, in future model deployment, the exact prediction window length may depend on the specific application of this framework. As we are primarily focused on model performance across a collection of contexts and activities, we proceed with a 30s window length for our analyses in this work.

\subsubsection{IMU Frame Size Sensitivity}
We further assess the impact of IMU frame size in conversation detection with the overall prediction window length fixed at 30 seconds. Towards this goal, we evaluate the inertial-only (CNN+Attention) and multimodal (Pure-Acoustic Model with CNN+Attention through Concatenation) frameworks on IMU frame sizes varying from 1 to 10 seconds in 1 second increments. We use a LOGO evaluation on the \textit{lab and semi-naturalistic} data and report the macro F1-score. 

Both inertial-only and acoustic-inertial models perform relatively consistently through the various IMU frame sizes (Figure \ref{Fig:imu-frame-sensitivity}). In the inertial-only model, model performance trends upward with larger frame lengths, achieving a maximum macro F1-score of 51.9$\pm$2.6\% at a frame size of 9s. In the acoustic-inertial model, model performance oscillates across IMU frame lengths and reaches a maximum macro F1-score of 77.2$\pm$1.8\% at a frame length of 2s. As we are primarily focused on the multimodal model, we proceeded with a 2-second IMU frame length for this classification task.

\subsection{Multimodality Benefits by Context}
To further understand specific contexts that benefit most from additional non-verbal communication in conversation sensing, we evaluate the single-modality and multimodal classifiers by activity type on the combined \textit{lab and semi-naturalistic} dataset. The specific activity types we consider are: 1) regular conversation, 2) conversation while eating, 3) reading out loud, 4) watching videos, and 5) music in background. For each activity type, we evaluate using LOGO cross validation the top single-modality and multimodal classifiers on all data segments that contain the target activity context and repeat the process for each activity type. For instance, for \textit{music in background}, the evaluation dataset is all data segments across all groups of the study that contained background music. These activity types are not mutually exclusive, except between \textit{regular conversation} and \textit{conversation while eating}. 

In this context-based analysis, there are significantly more class imbalances in the evaluation datasets. For instance, there are likely to be fewer instances of the \textit{conversation} class while watching videos. Therefore, we report both macro and weighted F1 scores. Macro F1-score evaluates model performance independently of the class distributions in the evaluation set. However, this makes the metric sensitive to rare classes and can be significantly influenced by rare labels. The weighted F1-score addresses this by taking into consideration class distributions. 

Across all activity contexts, the joint acoustic-inertial modality improves upon performance (weighted F1-score) of any single modality classifier. The addition of the inertial modality is most beneficial to the \textit{music in background} context, increasing the absolute mean value of the weighted F1-score of the audio-only model by 10.9\%. The benefit of the inertial modality makes sense, as people often tap to the beat of the music or move in other ways that are distinct from other social activities. This finding especially can be leveraged for analyzing conversations in settings that often contain background music, such as dining at a restaurant or gatherings at a bar. Overall, this evaluation highlights the effectiveness of joint acoustic-inertial sensing compared to single-modality sensing across a variety of contexts around which conversations are commonly centered.

\begin{figure}\centering
\subfloat{\label{a}\includegraphics[width=.49\linewidth]{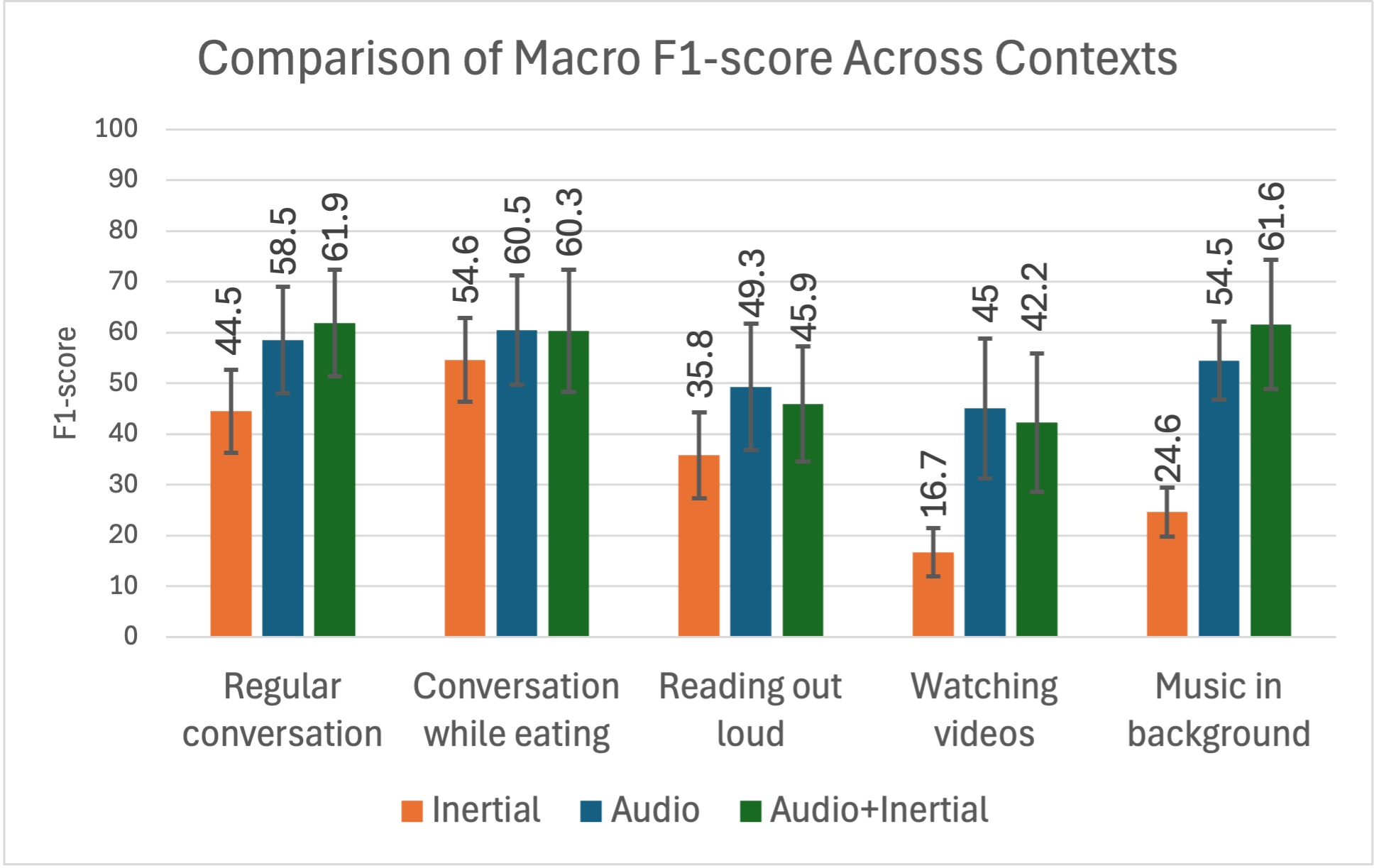}}
\subfloat{\label{b}\includegraphics[width=.49\linewidth]{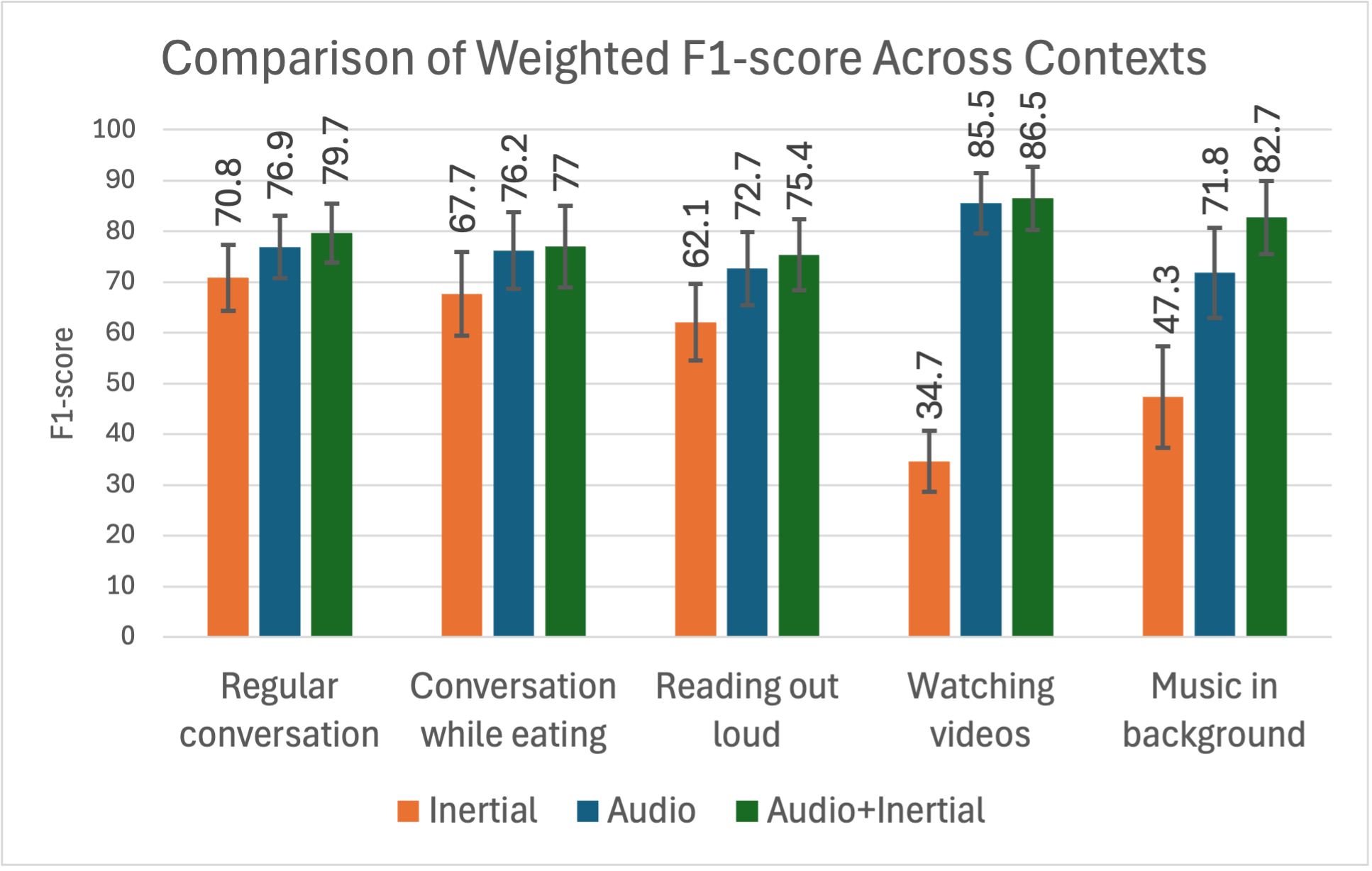}} 
\caption{Macro (left) and weighted (right) F1-scores across conversation} activity contexts.
\label{context_eval}
\vspace{-10pt}
\end{figure}

\begin{table}[htbp]
\small
\centering
\caption{Average macro-F1 score for audio and audio-inertial models with the audio sampled at three different sampling rates (16kHz, 2kHz, 1kHz).  L-LOGO: Training on all but one \textit{lab} sessions and evaluating on the holdout lab session. SN-LOGO: training on all but one session across the \textit{lab and semi-naturalistic} sessions and evaluating on the holdout session. SN: training on all \textit{lab} sessions and evaluating on all \textit{semi-naturalistic} sessions.}
\label{subsampled-summary}
\begin{tabular}{c c c c c c } 
%\begin{tabular}{@{}c *{5}{C{2cm}} @{}}

\toprule
Modality& Model& Frequency (kHz) & L-LOGO& L+SN-LOGO&SN\\
\hline
\multirow{3}{*}{Audio}
& \multirow{3}{*}{Pure-Acoustic Model } & 16 & 74.4 $\pm$ 3.3& 73.0 $\pm$ 2.0 &64.9 $\pm$ 2.8\\
& & 2 & 73.2 $\pm$ 3.4 & 74.1 $\pm$ 2.0 & 60.5 $\pm$ 1.6\\
& & 1 & 69.0 $\pm$ 2.3 & 72.3 $\pm$ 3.7  & 54.3 $\pm$ 2.5\\
\hline
 \multirow{6}{*}{Audio+Motion}
 & \multirow{3}{*}{\makecell{Pure-Acoustic Model \\ + SCNNB}} & 16 & 77.5 $\pm$ 3.0 & 78.0 $\pm$ 1.7 & 67.5 $\pm$ 2.5\\
 & & 2 & 69.7 $\pm$ 4.0 & 75.5 $\pm$ 2.0 & 64.5 $\pm$ 2.3 \\
 & & 1 & 71.9 $\pm$ 3.8 & 73.8 $\pm$ 1.8 & 61.7 $\pm$ 2.6\\
 \cline{2-6}
 & \multirow{3}{*}{\makecell{Pure-Acoustic Model \\ + CNN+Attention}} & 16 & 82.0 $\pm$ 3.0  & 77.2 $\pm$ 1.8 & 68.1 $\pm$ 2.8 \\
 & & 2 & 73.6 $\pm$ 3.2 & 75.7 $\pm$ 2.0 & 64.8 $\pm$ 2.7\\
 & & 1 & 73.2 $\pm$ 3.7 & 74.6 $\pm$ 1.8 & 62.2 $\pm$ 2.5\\
 \bottomrule
\end{tabular}
\vspace{-16pt}
\end{table}

\subsection{Audio Privacy Benefits of Multimodality}
\label{downsampled}
With the addition of inertial data to a traditionally audio-only classification problem, we explored whether information in the inertial data can reduce the amount of information required in the audio. The minimum sampling rate for intelligible speech is 16kHz, since most speech occurs below 8kHz \cite{schiel2012production}. Due to its intelligibility, 16kHz audio comes with privacy concerns about recording the content of speech. In contrast, inertial data sampled at 50Hz is less sensitive to privacy than audio data \cite{weiss2016actitracker}. Therefore, we investigated whether non-verbal communication captured through inertial data can supplement sub-sampled audio to increase privacy-preservation. %matic2013automatic

For this exploration, we created a sub-sampled audio dataset at 2kHz and 1kHz from the original audio dataset collected at 16kHz following the process described in section \ref{data_preprocessing}. We specifically selected 2kHz and 1kHz to isolate the frequency ranges where speech content becomes unintelligible and therefore represent increased audio privacy. As shown in previous work, sampling rates below 2kHz result in Word Error Rates (WER) exceeding 75\% and allows us to test if inertial data can compensate for the loss of audio information \cite{mollyn2022samosa}. We trained and evaluated the customized audio-only model and the top two combined audio-inertial models on audio sampled at 16kHz, 2kHz, and 1kHz. We report the macro F1-score for the same three evaluation setups described in section \ref{eval-setup}, which are: (1) evaluation in the \textit{lab}-only setting, (2) evaluation in the \textit{lab and semi-naturalistic} settings, and (3) evaluation in only the \textit{semi-naturalistic} settings using \textit{lab} trained models. 

As expected, the audio-only model's performance decreases across all three evaluation setups as the sampling rate decreases from 16kHz to 1kHz. In contrast, the top performing multimodal model (Pure-Acoustic Model and CNN+Attention) outperforms the audio-only model across all combinations of frequency and evaluation scenarios. While the audio-inertial model performance still decreases as audio quality decreases, the drop in model performance is not as significant as the drop in the audio-only model. Therefore, the addition of the inertial modality shows that IMU data can effectively supplement information lost in downgraded audio for detecting conversations. The combined audio and inertial framework is more robust to low-quality audio, and this finding can be leveraged to perform audio sensing at a lower sampling rate to maintain user privacy. %Most notably, multimodal model performance on the \textit{semi-naturalistic}-only evaluation decreases by only 5.9\% from the initial multimodal model's performance with audio at 16kHz compared to the 10.6\% decrease observed in the audio-only model. 

%The most significant decrease of 10.6\% of the original model performance is observed between 16kHz and 1kHz in evaluation in the \textit{semi-naturalistic}-only setting, where the model is trained on data collected in quiet settings and evaluated on data collected in noisier settings. 

\subsection{Cost-Benefit Analysis of Sampling Rates in Real-Time Implementation}
We conducted a cost-benefit analysis on model performance and smartwatch battery life of the Fossil Gen 5 as a function of audio and IMU sampling rates. Although for audio privacy measures we downsampled the audio to 1kHz, the Fossil smartwatches used in data collection are limited to 4kHz as the lowest microphone sampling rate. We compare the results of audio and IMU sampling rate combinations in Table \ref{tab:cost-benefit}. The additional IMU modality in the multimodal model provides a statistically significant improvement to the detection of conversations compared to the audio-only model. However, this benefit comes at a cost of 1.75 hours of reduced smartwatch battery life. Interestingly, we note that the model performs better with 4kHz audio compared to 16kHz audio. In listening to the audio downsampled to 4kHz, we hypothesize this is due to foreground voices still being audible and discernible but background speech becoming more distorted at 4kHz.

\begin{table}
    \centering
    \caption{Battery life and model performance across sampling rates. Battery is measured by the duration of data collection on the smartwatch using our data collection app until the battery is fully exhausted from one single, full charge. The IMU-only model is the SCNNB network, and the audio-IMU model is the Pure-Acoustic Model+SCNNB model. We report the F1-score of the models in the L+SN-LOGO evaluation scheme.}
    \renewcommand{\arraystretch}{0.9} % Reduces vertical space between rows
    \begin{tabular}{lcc}
    \toprule
    \textbf{Configuration} & \textbf{Battery} & \textbf{F1 Score} \\
    \midrule
    \textit{Audio 16 kHz} & & \\
    \hspace{3mm} + IMU 50 Hz & 3h 44m & 78.0 $\pm$ 1.7 \\
    \hspace{3mm} + IMU 25 Hz & 4h 18m & 75.9 $\pm$ 2.0 \\
    \hspace{3mm} + IMU 10 Hz & 5h 49m & 72.6 $\pm$ 2.2 \\
    \hspace{3mm} (Audio Only) & 5h 29m & 73.0 $\pm$ 2.0 \\
    \midrule
    \textit{Audio 4 kHz} & & \\
    \hspace{3mm} + IMU 50 Hz & 4h 55m & 78.0 $\pm$ 2.0 \\
    \hspace{3mm} + IMU 25 Hz & 5h 45m & 77.2 $\pm$ 1.8 \\
    \hspace{3mm} + IMU 10 Hz & 6h 40m & 76.5 $\pm$ 2.0 \\
    \hspace{3mm} (Audio Only) & 6h 54m & 74.6 $\pm$ 2.0 \\
    \midrule
    \textit{IMU Only (No Audio)} & & \\
    \hspace{3mm} IMU 50 Hz & 4h 26m & 53.6 $\pm$ 2.3 \\
    \hspace{3mm} IMU 25 Hz & 5h 28m & 52.9 $\pm$ 2.5 \\
    \hspace{3mm} IMU 10 Hz & 7h 06m & 49.6 $\pm$ 4.3 \\
    \bottomrule
    \end{tabular}
    \label{tab:cost-benefit}
    \vspace{-10pt}
\end{table}

\subsection{Validating the Challenge of Dynamic Environments for Conversation Detection}
%tread carefully
To better understand the model developed using our dataset in context with models developed on previous acoustic smartwatch datasets for conversations, we evaluated the trained model presented by Liang \textit{et al.} \cite{liang2023automated} on the audio in our collected dataset. Their study collected a \textit{semi-naturalistic} dataset with 32 hours of audio recorded in 18 homes while all household members engaged in a set of scripted activities and a \textit{free-living} dataset with 45 hours of audio recorded by 4 individuals in real-world settings without any activity constraints. Both datasets were recorded using smartwatch microphones as well. Notably, only their \textit{semi-naturalistic} dataset from home environments was used to train their acoustic model while both their \textit{semi-naturalistic} and \textit{free-living} datasets were used for model evaluation. On their \textit{semi-naturalistic} dataset, their model achieved a macro-F1 score of 76.2\% and on their \textit{free-living} dataset, they achieved a macro-F1 score of 89.2\%.

\begin{figure}
\includegraphics[width=12cm]{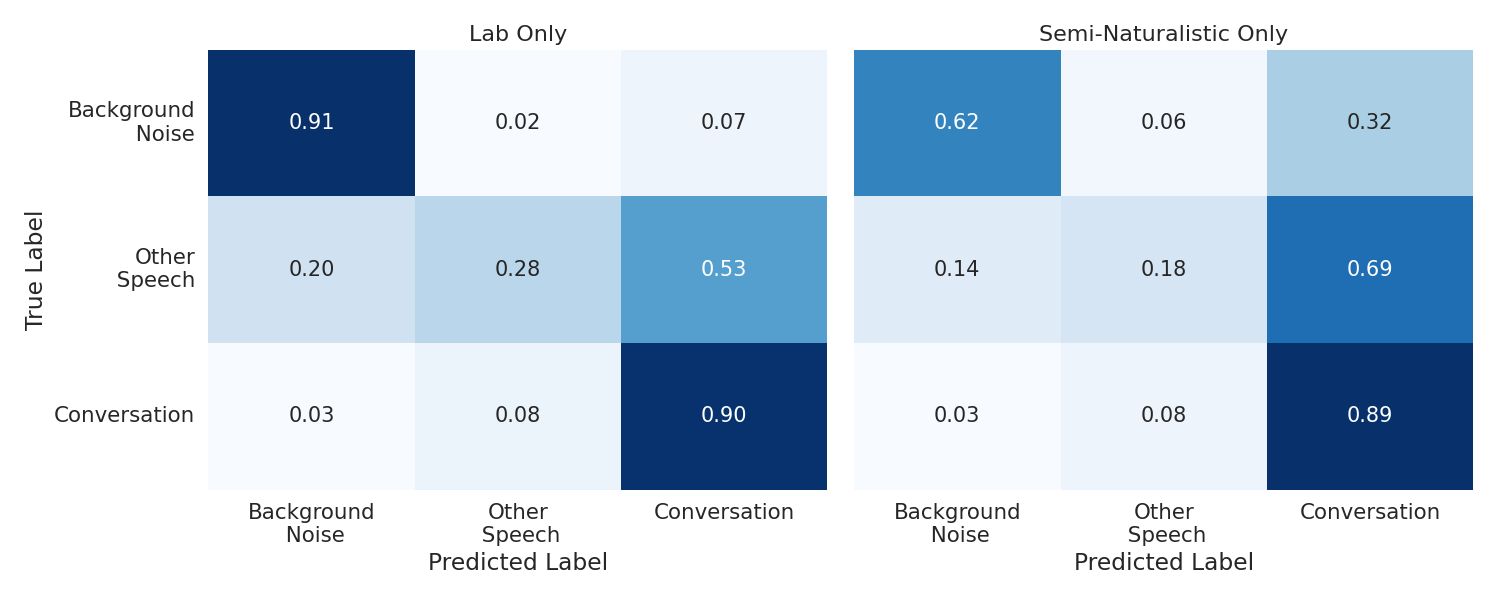}
\caption{Confusion matrices from inference of the pre-trained audio model on our dataset. Left: Inference on data from \textit{lab} setting (groups 1-4). Right: Inference on data from \textit{semi-naturalistic} setting (groups 5-11).}
\label{pre-trained-cm}
\vspace{-10pt}
\end{figure}

We leverage their model for inference only on both our \textit{lab}-only and \textit{semi-naturalistic}-only datasets. As seen in Figure \ref{pre-trained-cm}, their pre-trained model performs better on our \textit{lab} dataset (macro F1-score 68.7\%) than our \textit{semi-naturalistic} dataset (macro F1-score 52.3\%). This gap in performance between their datasets and our datasets emphasizes the acoustic difficulty of the environments in which we collected our datasets. 

Using their pre-trained model, the \textit{other speech} class is significantly confused with the \textit{conversation} class across both our \textit{lab} and \textit{semi-naturalistic} datasets. In our \textit{semi-naturalistic} dataset, conversations from passersby also resulted in \textit{background noise} confused with \textit{conversation}. Overall, this dataset comparison demonstrates a domain shift in the data where the training distribution (i.e., their \textit{semi-naturalistic} dataset from quieter home environments) differs from the test distribution (i.e., our \textit{semi-naturalistic} dataset from public, noisy environments). This highlights the uniqueness of our dataset and investigation of conversation detection in dynamic environments.

%comparison between these two datasets and shows the domain shift between these two datasets
%shows the necessity to train model directly on new dataset
%\textcolor{red}{also need to word it so that comparison to the macro-F1 score in the bar/restaurants doesn't detract from your work because your macro F1-score is even worse}

\subsection{Participant Handedness and Data Collection}

As previously discussed, to increase the ecological validity of the data collection study, participants were free to wear the smartwatch on either wrist. As seen in Table \ref{table:group_details}, a majority of participants were right-hand dominant and chose to wear the watch on their non-dominant left hand. However, during the study, it was observed that participants primarily gestured with their dominant hand, which aligns with previous studies on the relationship between gesturing and handedness \cite{nurcatak2018handedness}. Therefore, many participant gestures are not captured in the recorded inertial data. While the multimodal system is already an improvement from single-modality classifiers, inertial data from participants' dominant hand could help further clarify social activities, especially in noisy settings.

%\textbf{Model Size}

%As mentioned earlier, the future goal is to deploy the developed network onto edge devices for real-time inference of social interactions. The best performing acoustic and inertial model (CNN+Attention) have 763.2K and 2.8K parameters respectively. The multimodal model has 766.5K parameters

%% file: application.tex
Practical recognition of social interactions will enable a wide range of new applications. In this section, we further expand on applications of automatic conversation detection and discuss the extent to which our framework is suitable for these applications in light of our results. 
% discuss how our framework might be used in these scenarios.
% more sensitive applications (social diary), less sensitive applications (social isolation). A F1-score of 75% is suitable or not suitable Idk
% contextualize these applications in light of your results
 
\begin{itemize}
    \item \textbf{Loneliness and Social Isolation} Social isolation is as significant of a risk factor for health outcomes as traditional risk factors such as obesity \cite{pantell2013social}. After medical events such as experiencing a stroke, individuals' social networks decline and become less diverse \cite{hilari2016struggling}. Therefore, the proposed social interaction sensor can allow physicians and care providers to better support and understand the relationship between patients who have experienced such medical events and patient outcomes. 
    \item \textbf{Social Diary} Detecting social interactions allows individuals to maintain logs of their daily social interactions. The user can capture information, such as time and duration of their daily interactions, giving users a comprehensive view of their social activities. Longitudinally, these social diaries can increase speakers’ self-awareness of their social interactions and identify potential patterns of isolation.  %The analysis of gestures accompanying speech can give further insights into stress levels; as stress levels increase, a higher proportion of speech is accompanied by hand gestures \cite{lefter2016recognizing}. detect changes in social behavior, which can provide insights into individuals' health and wellness. The information in these diaries can also 
    %\item \textbf{Augmented Reality}  The ability to passively detect social interactions can enhance the immersive experience and interaction capabilities within augmented reality (AR) environments. With this social sensor, AR users can interact with virtual characters who respond to their gestures and verbal cues, resulting in more engaging and realistic interactions. 
\end{itemize}

At its current performance, our approach to sensing social interactions in noisy environments is suitable for these applications to a certain extent. To understand broad trends of loneliness and social isolation, the current framework could be acceptable. For other applications that require specific precision and recall to detect subtle nuances in conversation dynamics for enabling high-fidelity health analyses for instance, our system may need to be fine-tuned to the specific application context or improved with additional features discussed in the following section. Overall, though, our system represents a significant first step towards realizing these applications.
%For health-related analyses, the minimum acceptable accuracy of a medical artificial intelligence system is related to the average accuracy of human professionals performing the same task. Existing literature reports the minimum acceptable accuracy of such systems to be at least 80\%, as determined through comparison to commercial systems and physician surveys. However, this threshold depends on the specific medical task and the physicians who use the systems (e.g., specialists desire higher accuracy from medical AI systems compared to general practitioners) \cite{floares2010using, cabitza2020all}. 
%is related to the average accuracy of human professionals performing the same task, which therefore may set a higher accuracy requirement for conversation detection in medical settings

%% file: limitations_and_future_work.tex
While our work demonstrates the capabilities of smartwatch social sensing in naturalistic noisy settings, it is important to highlight its limitations and discuss future opportunities. First, though we evaluated our framework on a \textit{semi-naturalistic} dataset collected in real-world acoustic settings, we did not evaluate the framework on an \textit{in-the-wild} dataset where participants were unsupervised in their activities. In real-world settings, people can multitask, such as walking or driving, while engaged in a conversation. These activities that overlap with conversations, especially these activities that also have accompanying hand, arm, or wrist gestures, may create confusion with conversation-related gestures. Secondly, all participants in our data collection study were between the ages of 20-30 years old. Additional participants with more diversity in age would enhance the external validity of our results, since non-verbal and verbal communication vary across ages \cite{feldman2006factoring}. 
%\textcolor{red}{Additionally, conversations in real world settings are dynamic, and an adaptive window may be better suited for real world use of this system.}

Additionally, while speech processing tools alone are not currently suitable for social interaction analysis, they can extract information that can further assist in characterizing conversations. For instance, \textit{pyannote.audio} can perform speaker diarization \cite{Bredin23}. By segmenting audio according to who spoke when, the information gained could improve classification of the \textit{conversation} and \textit{other speech} classes. However, many speech processing tools have primarily been developed on datasets from controlled environments such as LibriSpeech \cite{panayotov2015librispeech} and Switchboard \cite{godfrey1992switchboard}, which come from audiobooks and telephone calls respectively. Therefore, these tools alone have limited applicability to detecting in-person conversations in acoustically challenging environments. However, coupled with our framework, these tools can contribute additional analyses.

Lastly, our current framework utilizes late fusion of the audio and inertial modalities to accommodate their differing sampling rates and preproccessing pipelines. However, existing research suggests that speech and gestures are tightly intertwined. Speech-related gestures exhibit temporal synchrony with produced speech, especially when there is a semantic relationship between the speech and accompanying gestures \cite{PouwPaxtonHarrisonDixon2020Acoustic, BergmannAksuKopp2011SpeechGesture}. Therefore, future work should explore alignment and resampling techniques to facilitate early fusion architectures. By processing raw audio and inertial signals jointly at the input level, we can better capture fine-grained relationships between speech and gestures, thereby potentially improving detection sensitivity.

%% file: conclusion.tex
We present the first joint acoustic-inertial sensing framework using off-the-shelf smartwatches for recognizing conversations. We demonstrate the benefits of inertial data in capturing non-verbal behaviors during in-person communication to aid acoustic sensing. To validate this framework, we collected two datasets: (1) a \textit{lab} dataset with 11 participants and (2) a \textit{semi-naturalistic} dataset with 24 participants performing the same group activities in acoustically challenging environments. Through a broad set of evaluations, we show the advantages of multimodal sensing for conversation detection in acoustically-challenging environments, which has been previously unexplored. Furthermore, we demonstrate the advantages of inertial data in aiding model performance across activity contexts and low-quality audio. This work advances the development of high-performing conversation detection systems by utilizing everyday wrist-worn devices to analyze both acoustic and inertial data. Lastly, we demonstrate a real-time implementation of our framework on commodity smartwatches, opening the door for building future systems with more fine-grained analyses of social dynamics for applications towards individual well-being, organizational behavior and more.